\documentclass[conference]{IEEEtran}
\usepackage{caption} 
\usepackage[utf8]{inputenc}

\usepackage{type1cm}
\usepackage{eso-pic}
\usepackage{color}
\PassOptionsToPackage{bookmarks=false}{hyperref}

\usepackage{xspace}
\ifCLASSINFOpdf
  \usepackage[pdftex]{graphicx}
  \usepackage{graphbox}
  \usepackage{multirow}
\else
\fi

\usepackage{listings}
\usepackage{paralist}

\usepackage{datatool}

\lstnewenvironment{code}[1][]%
{
\noindent
\minipage{1.0 \linewidth}
\vspace{0.5\baselineskip}
\lstset{
    language=Python,
    frame=single,
    captionpos=b,
    stringstyle=\ttfamily,
    basicstyle=\scriptsize\ttfamily,
    showstringspaces=false,#1}
}
{\endminipage}
\usepackage{footnote}
\usepackage{url}
\usepackage[T1]{fontenc}

\makeatletter
\newcommand\footnoteref[1]{\protected@xdef\@thefnmark{\ref{#1}}\@footnotemark}
\makeatother

\makeatletter
\def\ps@IEEEtitlepagestyle{
  \def\@oddfoot{\mycopyrightnotice}
  \def\@evenfoot{\mycopyrightnotice}
}
\def\mycopyrightnotice{
  {\footnotesize
  \begin{minipage}{\textwidth}
  \centering
  Copyright~\copyright~2017 IEEE.\\ 
  DOI: \url{https://doi.org/10.1109/BigData.2016.7841045}
  \end{minipage}
  }
}

\usepackage[
	pdfauthor={Andre Luckow et.\,al.},
	pdftitle={Automotive Deep Learning},
	pdfsubject={},
	pdfkeywords={},
	pdfdisplaydoctitle=true,
	plainpages=false,
	hypertexnames=false,
	hyperindex=false,
	bookmarks=false,
	colorlinks=false,
	pdfborder={0 0 0},
	linkbordercolor={0 0 0},
	pdftex
]{hyperref}

\begin{document}
\title{Deep Learning in the Automotive Industry: Applications and Tools \vspace{-4mm}}
\author{\small Andre Luckow$^{*}$, Matthew Cook$^{*}$, Nathan Ashcraft$^{\ddag}$, Edwin Weill$^{\dag}$, Emil Djerekarov$^{*}$, Bennie Vorster$^{*}$ \\
{\small $~^{*}$}BMW Group, IT Research Center, Information Management Americas, Greenville, SC 29607, USA\\
{\small $~^{\dag}$}Clemson University, Clemson, South Carolina, USA\\
{\small $~^{\ddag}$}University of Cincinnati, Cincinnati, Ohio, USA\\
\vspace{-8mm}}
\date{}
\maketitle

\begin{abstract}

Deep Learning refers to a set of machine learning techniques that utilize
neural networks with many hidden layers for tasks, such as image
classification, speech recognition, language understanding. Deep learning has
been proven to be very effective in these domains and is pervasively used by
many Internet services. In this paper, we describe different
automotive uses cases for deep learning in particular in the domain of computer
vision. We surveys the current state-of-the-art in libraries, tools and
infrastructures (e.\,g.\ GPUs and clouds) for implementing, training and
deploying deep neural networks. We particularly focus on convolutional neural
networks and computer vision use cases, such as the visual inspection process
in manufacturing plants and the analysis of social media data. To train neural
networks, curated and labeled datasets are essential. In particular, both the
availability and scope of such datasets is typically very limited. A main
contribution of this paper is the creation of an automotive dataset, that
allows us to learn and automatically recognize different vehicle properties. We
describe an end-to-end deep learning application utilizing a mobile app for
data collection and process support, and an Amazon-based cloud backend for
storage and training. For training we evaluate the use of cloud and on-premises
infrastructures (including multiple GPUs) in conjunction with different neural
network architectures and frameworks. We assess both the training times as well
as the accuracy of the classifier. Finally, we demonstrate the effectiveness of
the trained classifier in a real world setting during manufacturing process.

\end{abstract}

\bigskip
\begin{IEEEkeywords}Deep Learning, Cloud Computing, Automotive, Manufacturing \end{IEEEkeywords}

\section{Introduction}

Machine learning and deep learning has many potential applications in the
automotive domain both inside the vehicle, e.\,g.\ advanced driving assistance
systems (ADAS), autonomous driving,  and outside the vehicle, e.\,g. during 
development, manufacturing and sales \& aftersales processes. 
Machine learning is an essential component for use cases, such as predictive maintenance of vehicles, personalized infotainment and location-based services, business process automation, supply chain and price optimization. A common challenge of these applications is the need for storage and processing of large volumes of data as well as the necessity to deal with unstructured data (videos, images, text), e.\,g.\ from camera-based
sensors on the vehicle or machines in the manufacturing process. To effectively
utilize this kind of data, new methods, such as \emph{deep learning}, are required. 
Deep learning~\cite{Bengio-et-al-2015-Book,eosl} refers to a set of machine
learning algorithms that utilize large neural networks with many hidden layers
(also referred to as Deep Neural Networks (DNNs) for feature generation,
learning, classification and prediction.

Deep learning is extensively used by many online and mobile services,
such as the voice recognition and dialog systems of Siri, the Google Assistant, Amazon's Alexa and Microsoft Cortana, as well as the image
classification systems in Google Photo and Facebook. We believe that deep
learning has many applications within the automotive industry, such as computer
vision for autonomous driving and robotics, optimizations in the
manufacturing process (e.\,g.\ monitoring for quality issues), and connected
vehicle and infotainment services (e.\,g.\ voice recognition systems).

The landscape of infrastructure and tools for training and deploying deep
neural networks is evolving rapidly. In our previous work, we focused on
scalable Hadoop infrastructures for automotive applications supporting
workloads, such as ETL, SQL and machine learning algorithms for regression and
clustering analysis (e.\,g.\ KMeans, SVM and logistic
regression)~\cite{bd-workloads-infrastructure-2015}. While deep learning applications are similar to traditional big data systems, training and scaling of DNNs is challenging due to the large data and model sizes involved.
In contrast to simpler models, deep learning
involves millions, instead of hundreds, of parameters and larger datasets,
e.\,g.\ video, image or text data, for training. Training these models requires
scalable storage (e.\,g.\ HDFS), distributed processing, compute
capabilities (e.\,g.\ Spark), and accelerators (e.\,g.\ GPUs, FPGAs). Also, the
deployment of these models is a challenging task -- for deployment on mobile
devices the number of parameters and thus, the required amount of new input
data needs to be as small as possible. Modern convolutional neural networks
often require billions of operations for a single inference. 

\emph{This paper makes the following contributions: (i) It provides an
understanding of automotive deep learning applications and their requirements,
(ii) it surveys existing frameworks, tools and infrastructure for training DNNs
and provides a conceptual framework for understanding these, (iii) it provides
an understanding of the various trade-offs involved when designing, training
and deploying deep learning systems in different environments.} In this paper,
we demonstrate the usage of deep learning in two use cases implemented on cloud
and on-premise infrastructure, using different frameworks (Tensorflow, Caffe,
and Torch) and network architectures (AlexNet, GoogLeNet and Inception). We
show how to overcome various integration challenges to provide an end-to-end
deep learning enabled application: from data collection and labeling, network
training and model deployment in a mobile application. We demonstrate the 
effectiveness of the classifier by analyzing the classification performance of 
the mobile application during an extended test period.

This paper is structured as follows: in section~\ref{sec:use_cases}, we give an
overview of automotive use cases. We evaluate the current tools available for
deep learning in section~\ref{sec:tools}. We evaluate different deep learning 
use cases and models in conjunction with different public and proprietary 
datasets in section~\ref{sec:implementation_VisualInspection}.

\section{Automotive Use Cases}
\label{sec:use_cases}

Deep Learning techniques can be applied to many use cases in the
automotive industry. For example, computer vision is an area in which deep
learning systems have recently dramatically improved. Ng
et\,al.~\cite{2015arXiv150401716H} utilized convolutional neural networks for
vehicle and lane detection enabling the replacement of expensive
sensors (e.\,g.\ LIDAR) with cameras. Pomerleau~\cite{conf/nips/Pomerleau90}
used neural networks to automatically train a vehicle to drive by observing 
the input from a camera, a laser rangefinder and a real driver. In this section we describe a set of automotive use cases for deep learning.

\emph{Visual Inspection in Manufacturing:} The increased deployment of mobile
devices and IoT sensors, has led to a deluge of image and video data that is
often manually maintained using spreadsheets and folders. Deep learning can
help to organize this data and improve the data collection process.

\emph{Social Media Analytics:} Applications of computer
vision can extend to social media analytics. Consumer-produced image
data of vehicles made publicly available through social media can provide 
valuable information. Deep learning can assist and improve data
collection and analysis.

\emph{Autonomous Driving:} Different aspects of autonomous driving require
machine learning technologies, e.\,g\ the processing of the immense amounts of
sensor data (camera-based sensors, Lidar) and the learning of driving
situations and driver behavior.

\emph{Robots and Smart Machines:} Robotics requires sophisticated computer
vision sub-systems. Deep learning performs well for recognizing features in
camera images and other kinds of sensors needed to control the machine. While
object detection using DNN is well understood, a more challenging task in this
domain is object tracking. Further, deep learning enables self-learning robots
that become more intelligent over their lifetime.

\emph{Conversational User Interfaces:} Our connected vehicle already is the
platform for a large number of services. Voice dialog systems will become more
natural and interactive with deep learning allowing a hands-free interaction
with the vehicle.

In the following, we focus on the visual inspection application as an example to
understand the trade-offs between different datasets, model architectures,
training and scoring performance. Further, we analyze a use case in marketing
analytics to discuss performance in a real-world scenario.

\section{Background, Tools and Infrastructure}
\label{sec:tools}

In this section, we provide some background on deep learning and survey the 
landscape of tools for training neural networks.

\subsection{Background}

Neural networks are modeled after the human brain using multiple layers of
neurons -- each taking multiple inputs and generating an output -- to fit the
input to the output. The use of multiple layers of neurons allow the model to
learn complex, non-linear functions. These Deep Neural Networks (DNNs) are
particularly advantageous for unstructured data (which the majority of
data is) and complex, non-linear separable feature spaces. 
Schmidhuber~\cite{schmidhuber} provides an extensive survey of deep neural 
networks.

DNNs have shown superior results when compared to existing techniques for image
classifications~\cite{NIPS2012_4824}, language understanding, translation,
speech recognition~\cite{hinton2012deep}, and autonomous robots. Specialized
neural networks have emerged for different use cases, e.\,g.\ convolutional
neural networks (CNN), which pre-process and tile image regions for improved
image recognition.  Conversely, recurrent neural networks add a hidden layer that is
connected with itself for better speech recognition. Promising advances have
been made in automatically learning features (also referred to as
representation learning), through auto-encoders, sparse coders and other
techniques (see~\cite{DBLP:journals/corr/abs-1206-5538,HinSal06}). This is
particularly important as labeled data is difficult to obtain and the costs for
feature engineering are high.

There have been great advances in deep learning observable in the rapid
improvements of image classification accuracy in the ImageNet
competition~\cite{DBLP:journals/corr/RussakovskyDSKSMHKKBBF14}. The ImageNet
competition comprises a classification of a 1,000 category dataset of
$\sim$1.2 mio images. In 2015, the top~5 error rate achieved by a convolutional
neural networks (3.57\,\% for Microsoft's Residual Nets
approach~\cite{ms_deepresidlearning_2015arXiv}) was better than that of a
human (5.1\,\%). Another example is the recent success of
AlphaGo~\cite{alphago} in mastering the Go Game. Go is particularly challenging
as the search tree that needs to be mastered by the machine is very large: there are about 200 possibilities per move and a game consists of 150
moves leading to a search tree with a size of about $200^{150}$. AlphaGo uses an
ensemble of techniques, such a Monte-Carlo Tree search combined with a set of
deep neural networks.

\subsection{Deep Learning Libraries}

\begin{figure}[t]
  \centering
  \includegraphics[width=.49\textwidth]{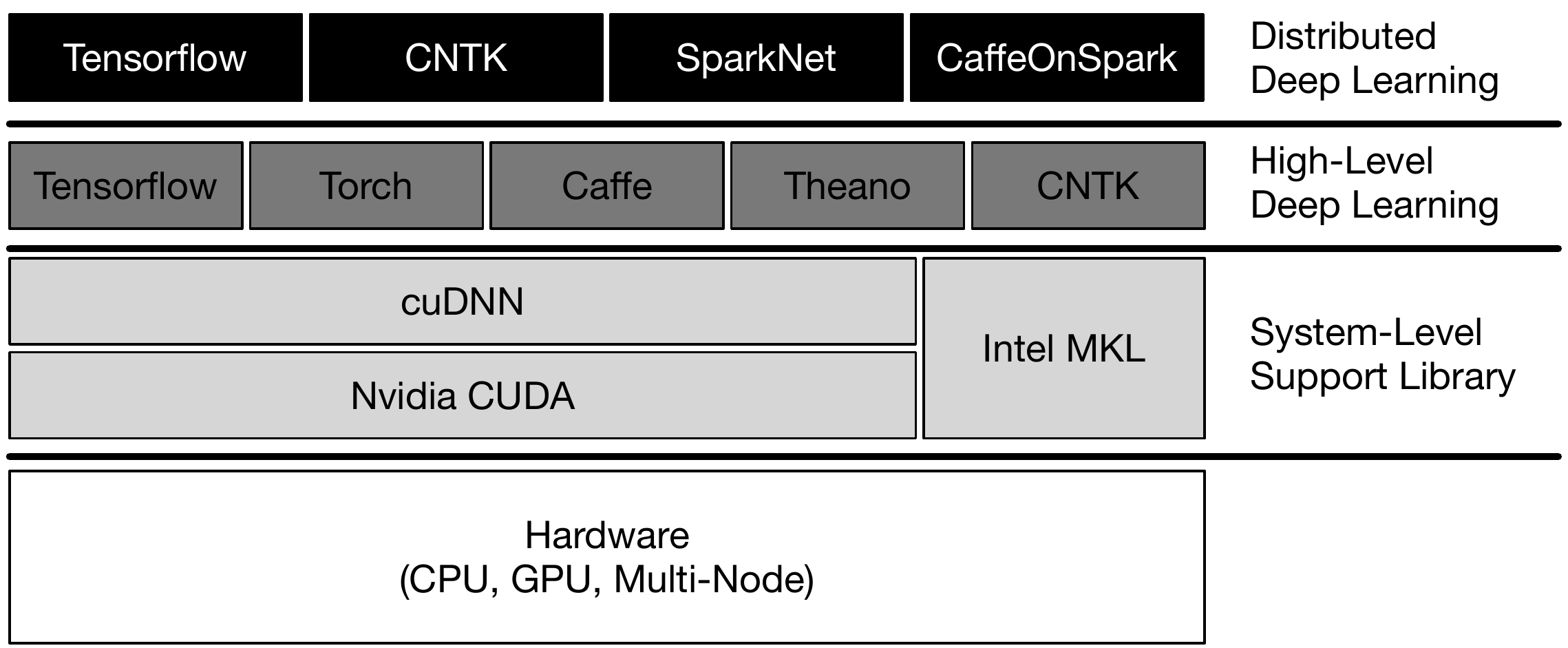}
  \caption{Deep Learning Software and Hardware }
  \label{fig:figures_deeplearning_stack}
\end{figure}

Neural networks -- in particular deep networks with many hidden layers -- are
challenging to scale. Also, the application/scoring against the model is more
compute intensive than other models.
Figure~\ref{fig:figures_deeplearning_stack} illustrates the different layers of
a deep learning system. GPUs have been proven to scale neural networks
particularly well, but have their limitations for larger image sizes. Several
libraries rely on GPUs for optimizing the training of neural
networks~\cite{DBLP:journals/corr/VasilacheJMCPL14}. Both NVIDIA's
cuDNN~\cite{cudnn} and Intel's MKL~\cite{dl_mkl} optimize critical deep
learning operations, e.\,g., convolutions. On top of these several high-level
frameworks emerged - some of which provide integrated support for distributed
training, while others rely on other distributed runtime engines for this
purpose.

Several higher-level deep learning libraries for different languages emerged:
Python/scikit-learn~\cite{scikit-learn},
Python/Pylearn2/The\-ano~\cite{pylearn2_arxiv_2013}, Python/Dato~\cite{dato},
Java/DL4J~\cite{dl4j}, R/neuralnet~\cite{RJournal_2010-1_Guenther+Fritsch},
Caffe~\cite{DBLP:journals/corr/JiaSDKLGGD14},
Tensorflow~\cite{tensorflow2015-whitepaper}, Microsoft
CNTK~\cite{cntk}, Amazon DSSTNE~\cite{dsstne}, 
MXNet~\cite{DBLP:journals/corr/ChenLLLWWXXZZ15},
Lua/Torch~\cite{collobert2011torch7} and Baidu's PaddlePaddle~\cite{paddle}. 
The ability to customize training and
model parameters differs; while some tools (e.\,g., DIGITS~\cite{digits},
Pylearn) focus on a high-level, easy-to-use abstractions for deep learning,
frameworks such as Theano and Tensorflow 
customizable low-level primitives. Further, several high-level frameworks
emerged: Keras~\cite{keras} provides a unified abstraction for specifying deep
learning networks agnostic of the backend. Currently, two backends: Theano and
Tensorflow are supported. Lasagne~\cite{lasagne} is another example for a
Theano-based library.

\subsection{Distributed Deep Learning}

The ability to scale neural networks -- i.\,e. to utilize networks with many
hidden layers and the ability to it train large datasets -- is critical in
order to train networks on large datasets in short amounts of time (important
to ensure fast research cycles). Neural networks utilizing millions of
parameters are generally more compute-intensive than other learning techniques.
The deeper the network, the higher the number of parameters and thus, the
larger the size of the model. In distributed approaches this model needs to be
synchronized across all nodes. To scale neural networks, the usage of
GPUs~\cite{cudnn}, FPGAs~\cite{fpga-dnn}, multicore machines and distributed
clusters (e.\,g.\ DistBelief~\cite{40565},
Baidu~\cite{DBLP:journals/corr/WuYSDS15}) have been proposed. In the following,
we particularly focus on approaches for supporting distributed GPU clusters.

Training large datasets on large deep learning models requires distributed
training, i.\,e.\ the usage of a cluster comprising of multiple compute nodes
n. Distributed machine learning requires the careful management of
computation and communication phases as well as distributed coordination. In
general, there are two types of parallelism to exploit: (i) data parallelism
and (ii) model parallelism (see Xing et al.~\cite{Xing:179} for a  
overview). Data parallelism is generally well-understood and easier to 
implement ; model parallelism requires the careful consideration of 
dependencies between the model parameters.

Most distributed deep learning libraries provide a distributed implementation
of gradient descent optimized for parallel learning. Implementing data
parallelism for gradient descent is well-understood: the data is partitioned
among all workers, which each computes parameter updates for its 
partition. After each iteration parameters are globally aggregated and the model is updated. Systems typically differ
in the way the model is stored and updated, and on how coordination between the workers is carried out. Some systems store the model centrally using a central master node, a set of nodes or dedicated parameter servers node(s), while others
replicate/partition the model across the worker nodes. Model updates can be
done synchronously or asynchronously (Hogwild~\cite{2011arXiv1106.5730N}).

\begin{table}[t]
	\begin{tabular}{|p{1.6cm}|p{1.3cm}|p{1.3cm}|p{1.3cm}|p{1.4cm}|}
	\hline
					&\textbf{CaffeOn\-Spark} &\textbf{SparkNet} &\textbf{Tensor\-flow} &\textbf{CNTK}\\ \hline
	Base Framework    &Caffe &Caffe &Tensorflow &CNTK \\ \hline
	Model Distribution     &replicated      &central (spark master)    &central (parameter server)    &replicated/ partitioned \\ \hline
	Model Update      &synchronous      &synchronous/ asynchronous      &synchronous &synchronous/ asynchronous (1 Bit SGD)  \\ \hline
	Communi\-cation   &MPI   &Spark &gRPC   &MPI \\ \hline
	\end{tabular}
	\caption{Distributed Deep Learning \label{tab:distributed_dnn}}
\end{table}

Hadoop~\cite{hadoop} and Spark~\cite{Zaharia:2010:SCC:1863103.1863113} emerged as de-facto-standard for data-parallel
applications~\cite{bd-workloads-infrastructure-2015}. However, support for deep
neural networks is still in its infancy. Spark provides a good platform for
data pre-processing, hyper-parameter tuning, and for distributed communication
and coordination. There is ongoing work to implement artificial neural networks
in Spark~\cite{spark_mlp} as part of its MLlib machine learning
library~\cite{mllib}. In addition, various approaches for integrating Spark
with frameworks, such as Caffe and Tensorflow emerged (see
table~\ref{tab:distributed_dnn}).

CaffeOnSpark~\cite{caffeOnSpark} provides several integration points with
Spark: it provides Hadoop InputFormats for existing Caffe formats, e.\,g.\ LMDB
datasets, and allows the integration of Caffe learning/training stages into
Spark-based data pipelines. CaffeOnSpark implements a distributed gradient 
descent. Gradient updates are exchanged using a MPI AllReduce across all 
machines.

SparkNet~\cite{2015arXiv151106051M} utilizes mini-batch parallelization to
compute the gradient on RDD-local data on worker-level. In each iteration, the
Spark master collects all computed gradients, averages them and broadcasts the
new model parameters to all workers. Similarly, TensorSpark~\cite{tensorspark}
utilizes a parameter server approach to implement a ``DownpourSGD'' (see
DistBelief~\cite{40565}).

Both Tensorflow~\cite{tensorflow2015-whitepaper} and
CNTK~\cite{ms_deepresidlearning_2015arXiv} provide different distributed
optimizer implementations. Tensorflow offers a relatively low-level API to
implement data- and model parallelism using a parameter server with synchronous
respectively asynchronous model updates. Communication is implemented using
gRPC. CNTK offers several parallel SGD implementations, which can be configured 
for training a network. The 1-bit SGD~\cite{1_bit_sgd} reduces the amount of 
data for model updates significantly by quantizing the gradients to 1-bit. 
Communication in CTNK is carried out using MPI.

In addition to the frameworks described above, several other systems
exist: FireCaffe~\cite{DBLP:journals/corr/IandolaAMK15} is another framework
built on top of Caffe; \cite{DBLP:journals/corr/VishnuSD16} and~\cite{arimo_tf}
provide alternative distributed Tensorflow implementations.

\subsection{Cloud Services}

\begin{figure}[t]
  \centering
    \includegraphics[width=.51\textwidth]{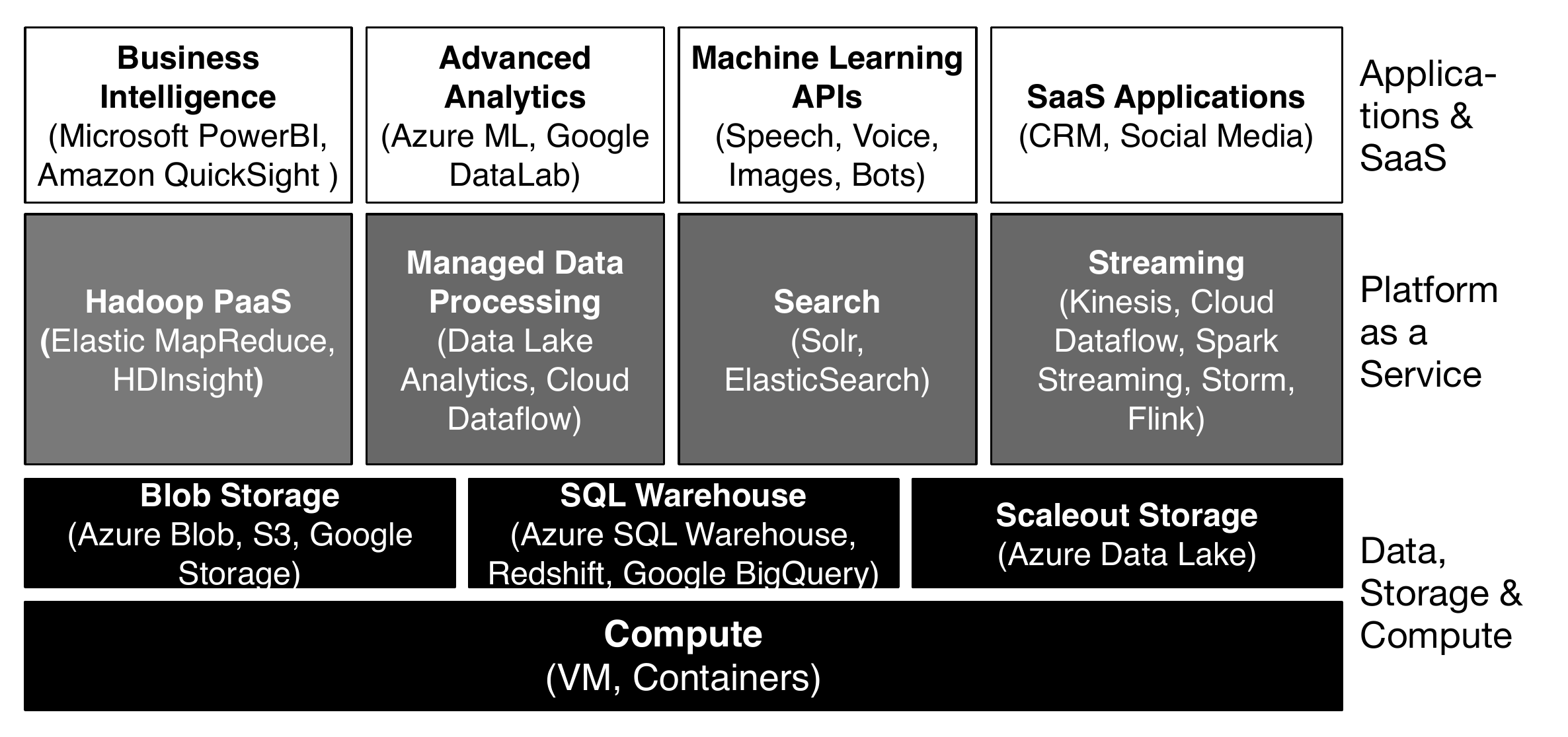}
  \caption{Cloud Infrastructure Layers}
  \label{fig:figures_cloud_dataanalytics}
\end{figure}

Cloud computing becomes increasingly a viable platform for implementing
end-to-end deep learning application providing comprehensive services for data
storage, processing as well as backend services for applications. In the
following we focus on data-related cloud services.
Figure~\ref{fig:figures_cloud_dataanalytics} categorizes services into three
layers: data storage, Platform-as-a-Services (PaaS) for Data and higher-level
Software-as-a-Service (SaaS).

An increasing number of infrastructure-as-a-service (IaaS) offerings with GPU
support exists: Amazon Web Services (AWS) provide the hardware necessary for
deep training and exploration while removing the necessity of obtaining a
physical system for computation. All services such as GPU computing and data
storage utilize the cloud and can therefore be managed accordingly. Amazon Web
Services Elastic Compute Cloud (EC2) is a service that provides cloud computing
with resizable compute capabilities including up to four K520 Grid
GPUs~\cite{amazon_gpu}. Similar capabilities have been announced by Microsoft. 
While Google does not provide GPU as part of its Google Compute Engine Service, 
it provides a managed PaaS environment for Tensorflow, which offers GPU 
support~\cite{google_cml}.

Every cloud provider provides a managed Hadoop/Spark environment. There are 
minor differences in the feature: Amazon Elastic MapReduce~\cite{emr} relies on 
his own Hadoop distribution and also supports Presto and Mapr, Microsoft's 
HDInsight~\cite{azure-hdi} is based on Hortonworks, Google's 
Dataproc~\cite{google-dataproc} also utilizes his own distribution. 
Typically, these Hadoop environments can read data from Blob storage and 
provide a HDFS cluster. They provide core nodes, which offer important services 
a such as the Namenode and YARN, and worker nodes, which can be scaled with 
demand.

Further, there are various cloud products related to search and streaming data.
Azure provides a native search engine: Azure Search that can easily index Azure
storage. Both Amazon and Microsoft provide a managed ElasticSearch environment.
Increasingly, there is the need to react on incoming data streaming using
various streaming tools and platforms. Topically, streaming systems consists of
a broker engine (e.\,g.\ Kafka) and processing tools on various levels (e.\,g.,
Storm and Spark Streaming). Azure offers support for Streaming via the Azure
Event Hub and Storm at the moment.

In addition several higher-level machine learning emerged. Google's Prediction
API~\cite{google_prediction} was one of the first services offering machine
learning classifications and predictions in the cloud. Microsoft's Azure
ML~\cite{azureml} and Amazon Machine Learning~\cite{amazonml} offer similar
services. These services allow simple and fast access to machine learning
capabilities. Models are easily deployed and published for further usage.
In particular, Google and Amazon often provide
black-box models with limited abilities for calibration of the model. Microsoft
allows the creation of more general data pipelines supporting custom R and
Python code.

A lot of shrink-wrapped solutions that offer deep learning capabilities behind
a high-level cloud API (Platform as a Service), e.\,g.\ for
advanced machine learning tasks, such as facial recognition, computer vision and
machine translation, are often based on deep learning. Examples are Microsoft's
Project Oxford~\cite{project_oxford}, Google's Vision API~\cite{google_vision} 
and Natural Language API~\cite{google_nlp}, and IBM's Watson developer cloud 
(AlchemyVision API)~\cite{ibm_watson}. The
core of these services relies on deep learning technologies. However, these
services are constrained by the number of categories they support -- Project
Oxford's Image API supports only 86 categories. Also, training on
custom categories and data, via transfer learning, is often not possible.

\begin{table}[t]
	\centering
\begin{tabular}{|p{1.5cm}|p{1.8cm}|p{2cm}|p{2.2cm}|}
	\hline
	&\textbf{Amazon} &\textbf{Microsoft} &\textbf{Google} \\ \hline
	PaaS APIs &Amazon Rekognition &Project Oxford &Prediction API, Google Vision 
	API, Speech API, Natural Language \\ \hline
	Advanced Analytics &Amazon Machine Learning &Azure ML (incl. Jupyter 
	Notebooks) &Cloud Machine Learning (with GPUs), DataLab (Jupyter 
	Notebooks) \\ \hline
	Deep Learning Framework &DSSTNE, MXNet &CTNK &Tensorflow\\ \hline
	Data Platform as a Service &Elastic MapReduce &HDInsight, Data Lake 
			Storage/Analytics &Google Dataproc, Cloud Dataflow\\ \hline
	
	Data Storage &S3, Redshift &Azure Storage, SQL Datawarehouse &Cloud Big Table\\ \hline
	Compute Nodes &EC2 (with GPU) &Azure Compute (GPU announced) &Google Compute Engine (no GPU)\\ \hline
\end{tabular}
\caption{Cloud Services for Data Analytics}
\end{table}

\section{Implementation and Evaluation}
\label{sec:implementation_VisualInspection}

In this section, we evaluate different convolutional neural networks for object 
detection on two different datasets (i) images collected at a manufacturing 
facility and (ii) a hand-curated social media datasets. Further, we evaluate 
different deep learning frameworks to understand training and inference 
performance.

\subsection{Experiments and Evaluation}

In the following, we evaluate different frameworks for training the deep neural
networks. For experiments, we use a machine with 2 CPUs, a total of 8 cores,
128\,GB memory and a TITAN X GPU. Further, we utilize Amazon Web Services GPU nodes (g2.8xlarge), which provides 32 cores, 60\,GB memory and 4 K520 GPUs~\cite{amazon_gpu}.
For training the Caffe and Torch models, we use DIGITS~\cite{digits} and the models provided with it. For Tensorflow, we
adapted the provided AlexNet implementation~\cite{alexnet_tf}.

\subsection{Datasets}

We identified a set of datasets relevant for the automotive industry (see
Table~\ref{tab:datasets}). ImageNet is one of the largest publicly available
datasets. The usage of ImageNet and transfer learning is particularly suited
for social media analytics and other forms of web data analysis. For enterprise
use cases it is required to curate custom datasets. In particular for advanced
applications, such as autonomous driving, it is essential to create suitable
datasets, as datasets like Traffic Signs~\cite{Houben-IJCNN-2013},
Places~\cite{Places} and Kitti~\cite{Geiger2012CVPR}, are designed for
benchmarking primarily. Real-world applications require more data.

Further, we created a new dataset using data created during the visual 
inspection process. This dataset contains images from 4 vehicle types and 25 
camera perspectives, i.\,e.\ a total of 100 categories, that were captured 
using the mobile application described below. It currently consists of 
82,011\,images.

\begin{table}[t]
 \centering
\begin{tabular}{|p{1.6cm}|c|c|c|}
\hline
        &\textbf{Categories} &\textbf{Number Images} &\textbf{Size}\\
\hline
Visual Inspection &100 &82,011 &9\,GB (LMDB)\\
\hline
Cars~\cite{krause20133d}    &196 &16,185 &1.87\,GB (LMDB)\\
\hline
ImageNet 2012~\cite{DBLP:journals/corr/RussakovskyDSKSMHKKBBF14}  &1000 &1,281,167 &130\,GB (LMDB)\\
\hline
Traffic Signs~\cite{Houben-IJCNN-2013} &43 &1,200 &54.MB (LMDB)\\
\hline
Places~\cite{Places} &205 &2.5 mio &38.2\,GB (LMDB)\\
\hline

\end{tabular}
\caption{Object Detection Datasets \label{tab:datasets}}
\end{table}

\subsection{Visual Inspection for Manufacturing}

\begin{figure}[t]
  \centering
    \includegraphics[width=.49\textwidth]{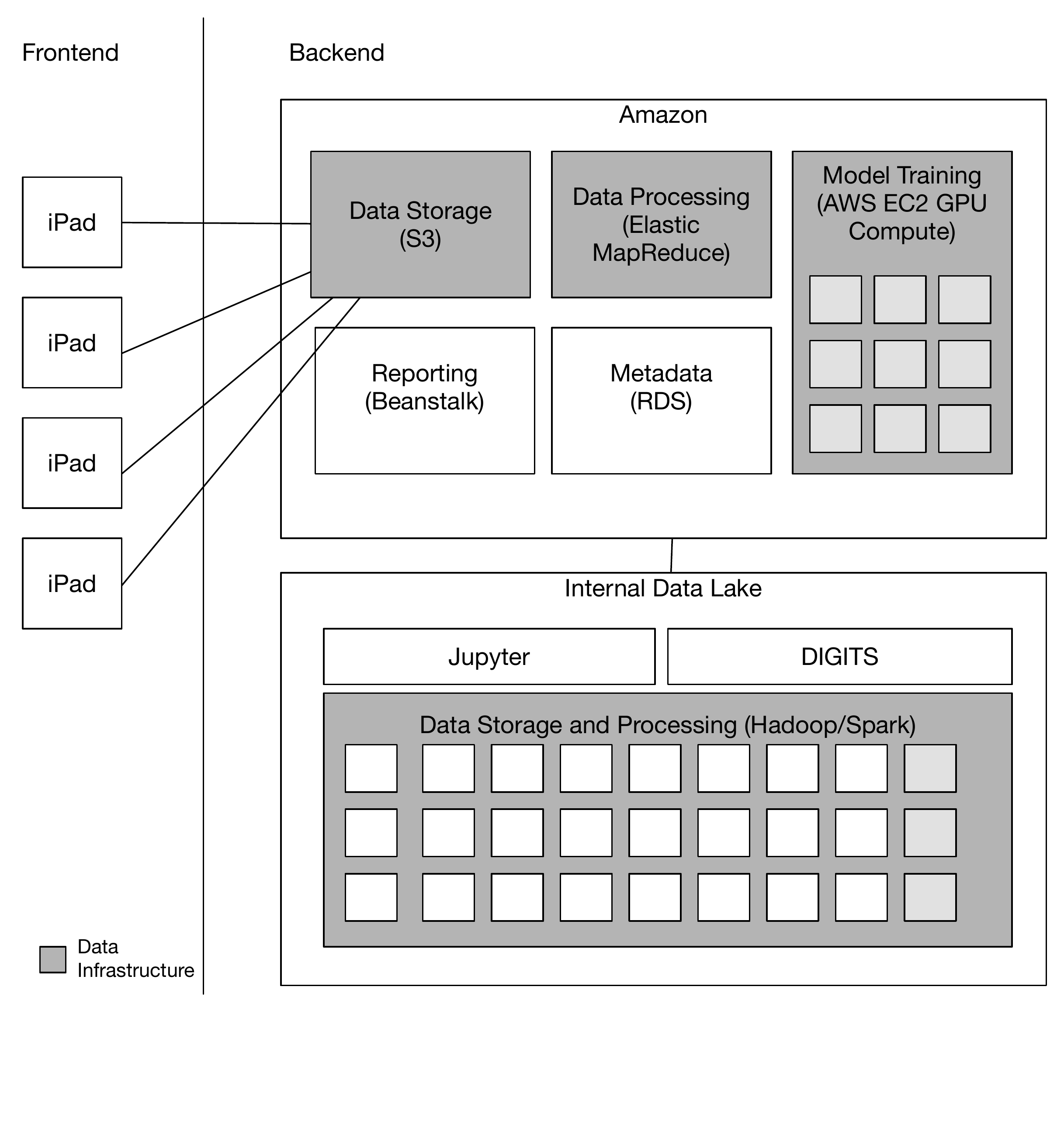}
  \caption{Visual Inspection Application Architecture}
  \label{fig:figures_q-gate}
\end{figure}

To support the visual inspection process during manufacturing and to aid
data collection, we built an iPad application. The application is used by 
associates to document a subset of produced vehicles using approximately 20 
walk-around pictures. Figure~\ref{fig:figures_q-gate} 
shows the architecture of the application and the deep learning backend. The 
iPad automatically uploads taken images to Amazon S3; The metadata is stored in 
a relational database backend. Both data movement and storage are encrypted. For
data-processing, we utilize a combination of Hadoop/Spark and GPU-based deep
learning frameworks deployed both on-premise and in the cloud. For data
pre-processing and structured queries, we rely on Hadoop and
Spark~\cite{luckow-2015}; for deep learning we rely on some GPU nodes.

The trained network is integrated into the iPad application to validate new
images taken by the associate. For this purpose, we compiled Caffe for iOS and
used the trained model files.

\subsubsection{Models Training}

We investigate different convolutional network architectures.
Table~\ref{tab:model} gives an overview of the different model
architectures investigated. In the following, we compare the AlexNet and
GoogLeNet architectures implemented on top of Tensorflow, Caffe and Torch.

\begin{table*}[t]
\centering
\begin{tabular}{|p{3cm}|p{3cm}|p{3cm}|p{3cm}|}
\hline
\textbf{Network}        &\textbf{Number Parameters} &\textbf{Number Layers} &\textbf{ImageNet Top~5 Error}\\
\hline
AlexNet (2012)~\cite{NIPS2012_4824} &60\,mio. &8~(5 convolutional, 3 fully~con\-nected) & 15.3\,\%\\
\hline
GoogLeNet (2014)~\cite{DBLP:journals/corr/SzegedyLJSRAEVR14, googlenet-v2}  &5\,mio. &22 &6.7\,\%\\
\hline
VGG (2014)~\cite{vgg_very_deep_conv} &$\sim$140\,mio. &19~(16 con\-volutio\-nal, 3 fully connected) &7.3\,\%\\
\hline
Inception v3  (2015)~\cite{google_inception_2015arXiv} &25\,mio. &42 &3.58\,\%\\
\hline
Deep Residual Learning (2015)~\cite{ms_deepresidlearning_2015arXiv} &$\sim$60\,mio.  &152 &3.57\,\%\\
\hline
\end{tabular}
\caption{Convolutional Neural Network Models \label{tab:model}}
\end{table*}

\begin{figure}[t]
  \centering
    \includegraphics[width=.5\textwidth]{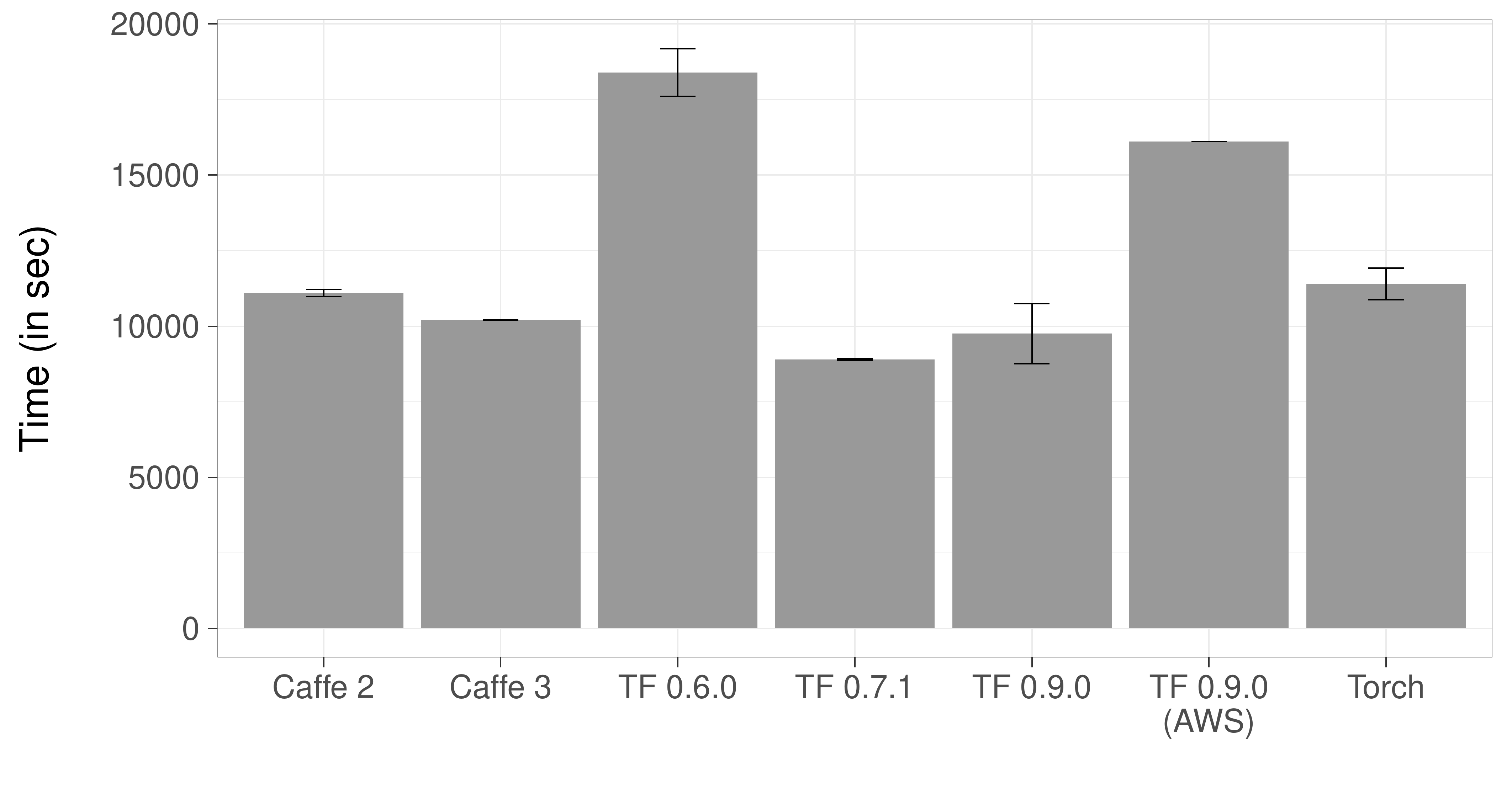}
  \caption{\textbf{Visual Inspection Training Times for AlexNet on Caffe, 
  Tensorflow (TF), and Torch:} With the maturation of the different frameworks 
  and the underlying system-level libraries (such as cuDNN), performance 
  improves significantly with newer framework versions. The GPU hardware is 
  another important consideration as seen is the performance on Amazon EC2 
  (AWS), which only provide older GPUs.}
  \label{fig:experiments_alexnet}
\end{figure}

Figure~\ref{fig:experiments_alexnet} illustrates the training times observed
for 30 epochs of the data with different frameworks.  There is an improvement in
the training times between Caffe 2 and 3 as well as TensorFlow 0.6 and 0.7.1.
This can be attributed to the usage of newer versions of cuDNN (v4).  We
achieved the best training time with Tensorflow 0.7.1.  TensorFlow 0.9.0 is also
evaluated as the breaking edge version of the software.  In our experiment, the
training time is slightly slower than with previous Tensorflow, which can be
attributed to a single factor; inconsistent training times per iteration.
With TensorFlow 0.7.1, each iteration has a standard deviation over all 30
epochs less than 2 seconds. Conversely, TensorFlow 0.9.0, while mostly
consistent, has a few iterations which cause the standard deviation to be much
larger.  This can be seen in figure~\ref{fig:experiments_alexnet} as the error
bar for TensorFlow 0.7.1 is small in comparison to its counterpart for
TensorFlow 0.9.0.  This inconsistency with some iteration times results in a
longer overall training time.

We also compare performance using TensorFlow 0.9.0 on a local machine versus a
machine utilizing cloud services. Figure~\ref{fig:experiments_alexnet} 
illustrates a performance
comparison of the EC2 web service and a local machine containing a TitanX
GPU. The local system utilizing TensorFlow provides a quicker training
time for the dataset
provided, however, AWS EC2 would be a great option if a physical machine with
dedicated hardward is unavailable as the training time is ~1.5x longer than
that of the local machine with the TitanX. The GPU used in AWS EC2 provides
the same amount of compute cores as the TitanX, however, the clock speed is
slower, allowing faster computation to occur on the TitanX. Also, the K520
GPU provides 8\,GB of device memory, while the TitanX provides 12\,GB allowing
for larger models or larger batch sizes to be used for computation.

Further, a software comparison is made between cuDNN v4 and v5.1 on the TitanX.
The update in software directly leads to decreased training time on the same
hardware from ~9,750 seconds to ~7,380 (a decrease of ~25\,\%). 
For larger datasets and larger networks, this update greatly improves training
time allowing for faster production of models.

\begin{figure}[t]
  \centering
    \includegraphics[width=.4\textwidth]{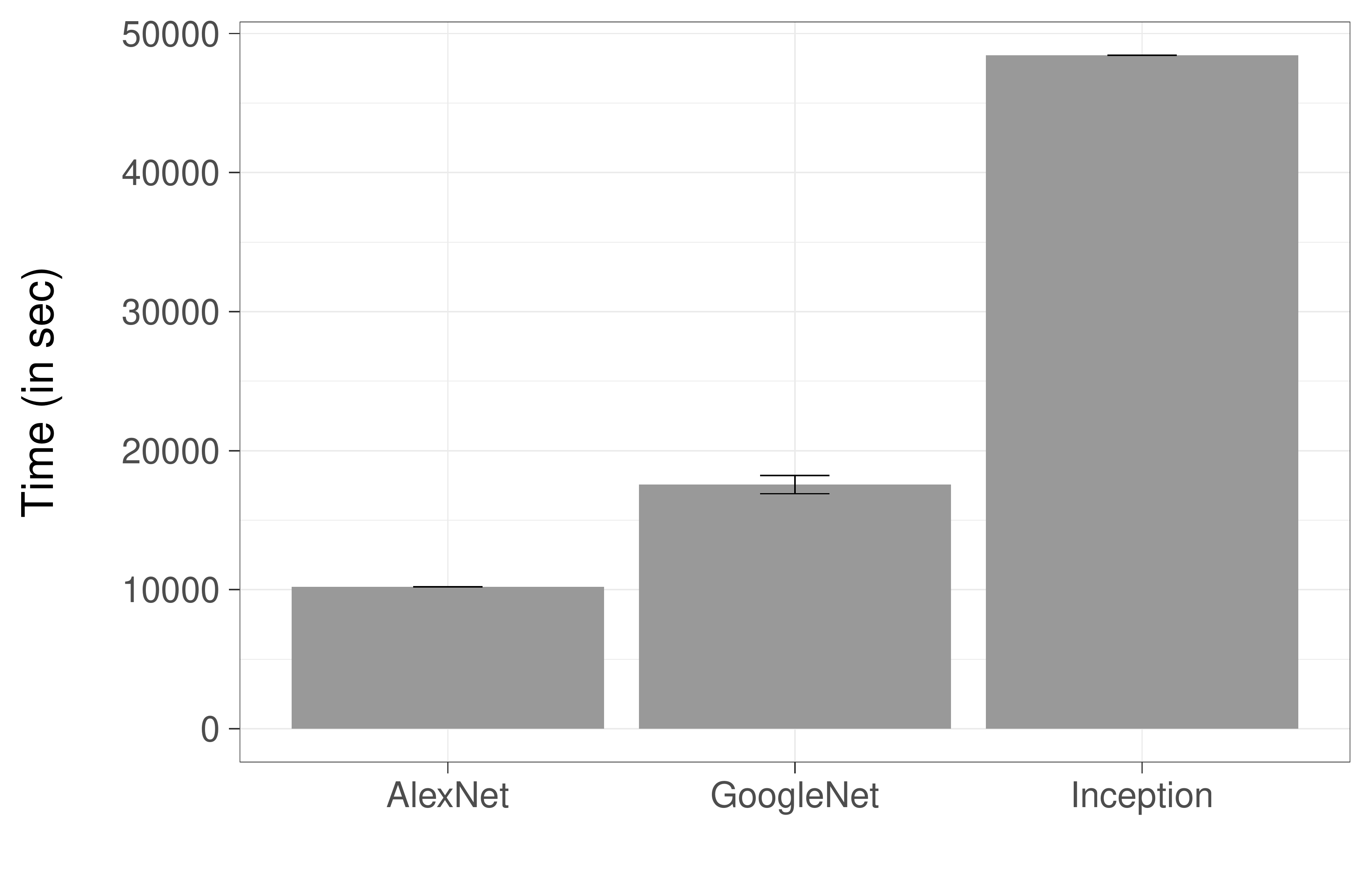}
  \caption{\textbf{Visual Inspection Dataset Training Times for AlexNet, GoogLeNet and Inception:} With the increased complexity of these networks the training times increase.}
  \label{fig:experiments_google_alexnet}
\end{figure}

Figure~\ref{fig:experiments_google_alexnet} compares the training times for
AlexNet and GoogLeNet using Caffe. Training GoogLeNet is 70\,\% slower than
AlexNet mainly due to the higher complexity of the networks (more deep layers).
Inception overshadows both AlexNet and GoogLeNet due to the complexity and deep
nature of the network.

\begin{figure}[t]
  \centering
    \includegraphics[width=.51\textwidth]{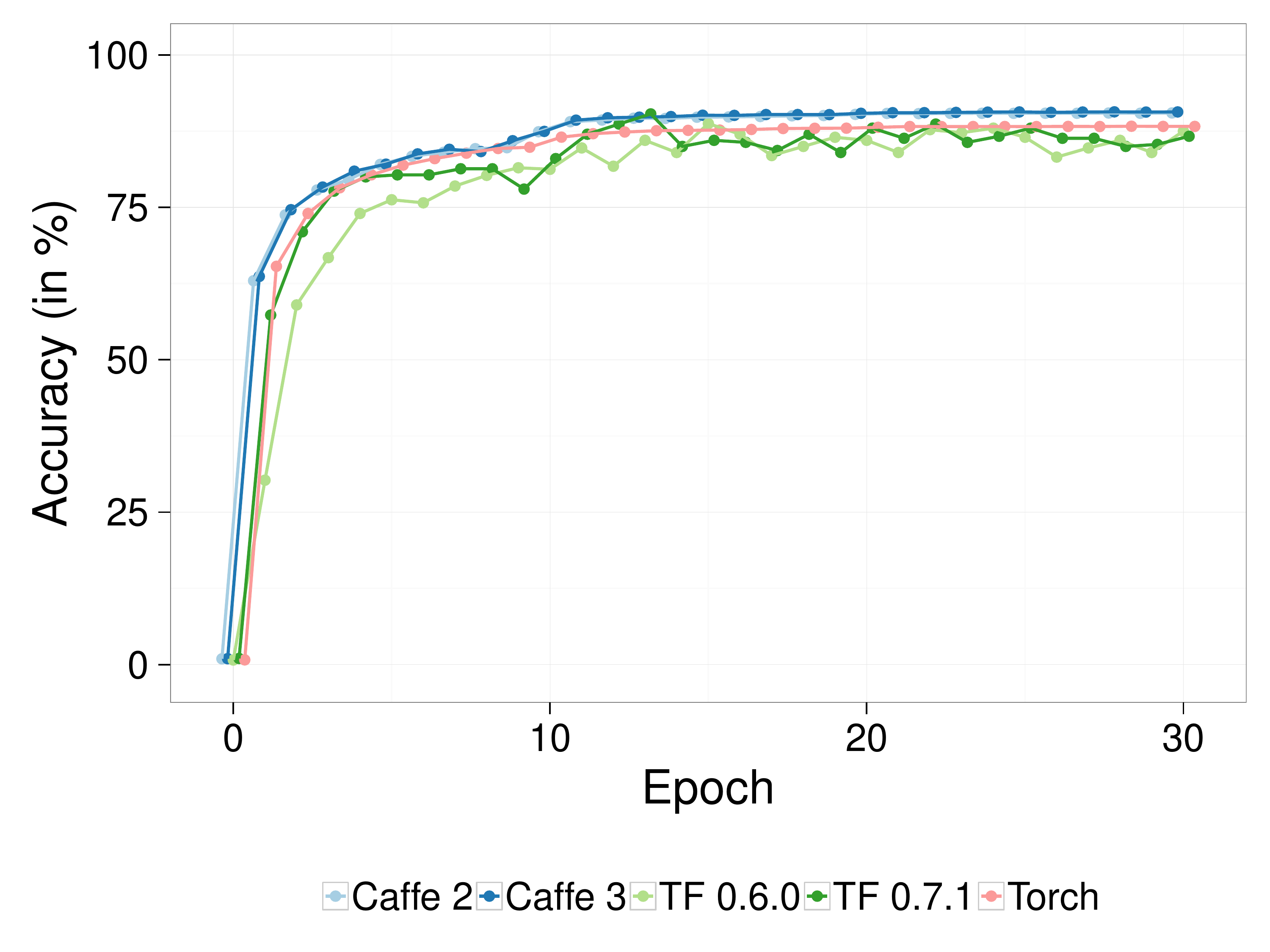}
  \caption{Visual Inspection Accuracies and Convergence for AlexNet on Caffe, Tensorflow (TF), and Torch}
  \label{fig:experiments_alexnet_peak_accuracy}
\end{figure}

Our investigation also included a comparison of the peak accuracies achieved from
training our models on different frameworks as well as the time in epochs it
took to reach them. Figure~\ref{fig:experiments_alexnet_peak_accuracy} shows
this comparison for the AlexNet model. There are no changes in peak accuracy
performance between versions of Caffe or Tensorflow. This is expected behavior
since only the underlying implementation of the frameworks, and not the
algorithm of the model, have been changed between versions. The best peak
accuracy we recorded was 94\,\% with all versions of Tensorflow.

Lastly, we compared the number of epochs required by each framework to achieve
its peak accuracy: TF shows the quickest convergences with 17 epochs in
average, followed by Torch with 23 epochs and Caffe with 28 epochs. Fewer
epochs directly translate to a shorter training time.

\subsubsection{Multiple GPU Training}
\label{sec:imagenet_training_multiple_gpu}

The ability to train CNNs on large datasets of images for recognition and
detection is critical. In the following we analyze the training time for the
Visual Inspection and ImageNet datasets in conjunction with multiple GPUs.
We utilize the ImageNet 2012 dataset consisting of 1,281,167 images and 1,000
classes, which is significantly larger than the Visual Inspection datasets with
~82,011 images. For training, we use the Caffe framework with GoogLeNet and
AlexNet for the Visual Inspection dataset.

\begin{figure}[t]
  \centering         \includegraphics[width=.47\textwidth]{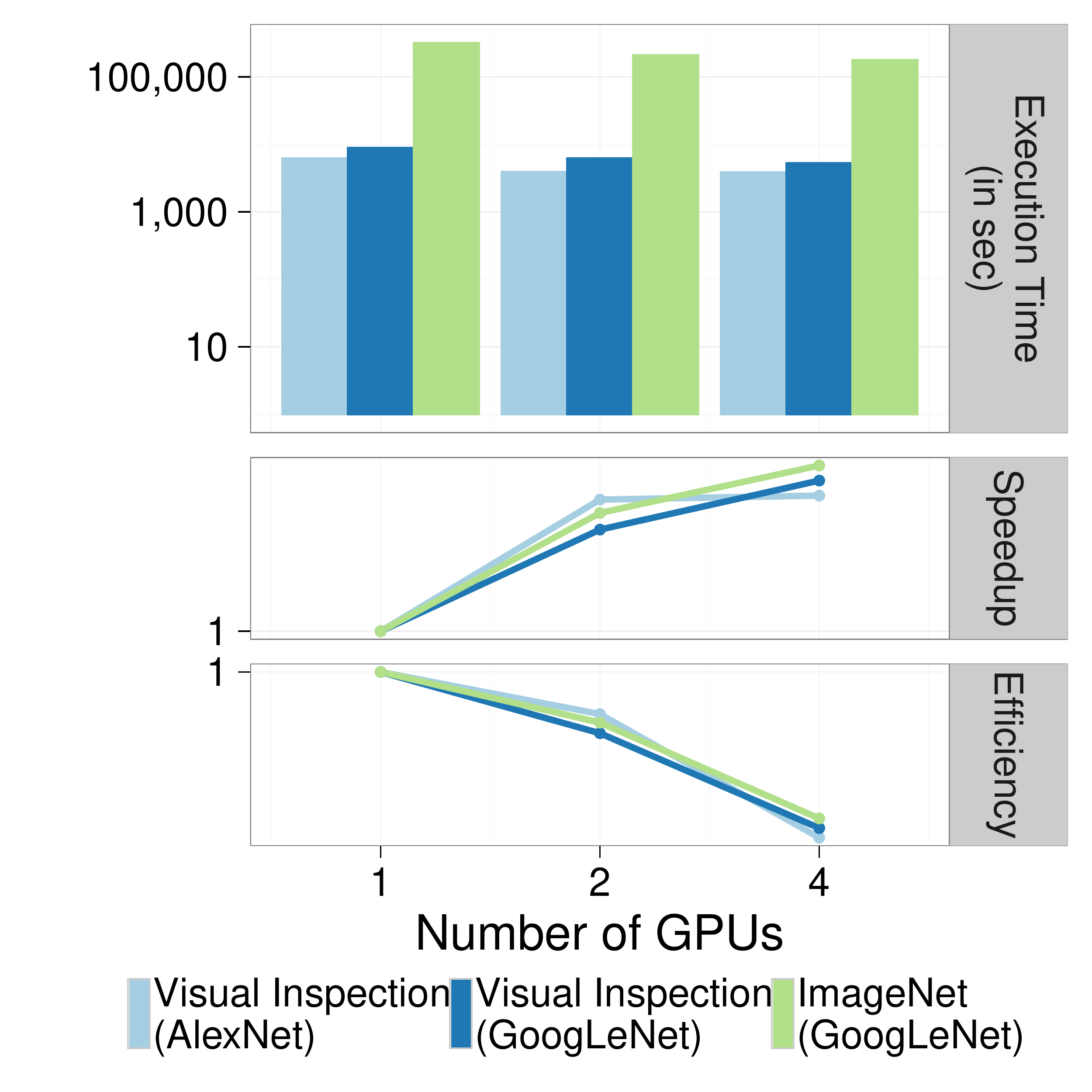}
  \caption{\textbf{ImageNet and Visual Inspection Training Times for 
  GoogLeNet/AlexNet on Multiple GPUs (Log Scale):} Multiple GPUs are particular 
  for large datasets advantages. For ImageNet we were able to observe a speedup 
  of 1.8 with 4 GPUs corresponding to an efficiency of 0.45. For the smaller 
  Visual inspection dataset the efficiency is slightly worse with 0.4. 
  GoogLeNet's training time is longer than AlexNet; efficiency is 
  better for GoogLeNet.}
  \label{fig:imagenet_multi_gpu_train}
\end{figure}

We are able to achieve similar accuracies for multiple GPUs training as for
single GPU, e.\,g., for ImageNet a top-5 accuracy of 87\,\% was obtained.
Figure~\ref{fig:imagenet_multi_gpu_train} illustrates the execution time,
speedup, and efficiency for up to 4 GPUs. As expected the training time
decreases with the number of GPUs. The efficiency, however, decreases
nonlinearly pointing out that even though the execution time is decreasing, the
addition of GPUs is causing an inefficiency. The speedup of using 2 GPUs is 1.5
which corresponds to an efficiency of 0.8, while training using 4 GPUs shows a
speedup of 1.8, corresponding to an efficiency of 0.45. For the significantly
smaller Visual Inspection dataset a maximum speedup of 1.6 corresponding to an
efficiency of 0.4 was observed with 4 GPUs. This shows that the use of more 
GPUs is not always advantageous as the efficiency drops quickly if the GPUs are 
not utilized fully.  Another interesting observation is the behaviour of 
GoogLeNet vs. AlexNet: while the training time for GoogLeNet is slightly 
higher, the scaling efficiency of GoogLeNet is slightly better than for AlexNet.

\subsubsection{Model Deployment}

For deployment of deep learning models in particular in mobile and embedded
environments, the performance is essential. The more complex the network, the
more compute-intensive the scoring process. There are two options for deploying
the model: (i) on the mobile device and (ii) in the backend system. An
important concern in particular for mobile deployment is the model size, which
depends on the number of parameters in the model. The trained GoogLeNet model
is about 43\,MB in size, while the AlexNet model is 230\,MB.

\begin{figure}[t]
  \centering
  \includegraphics[width=.4\textwidth]{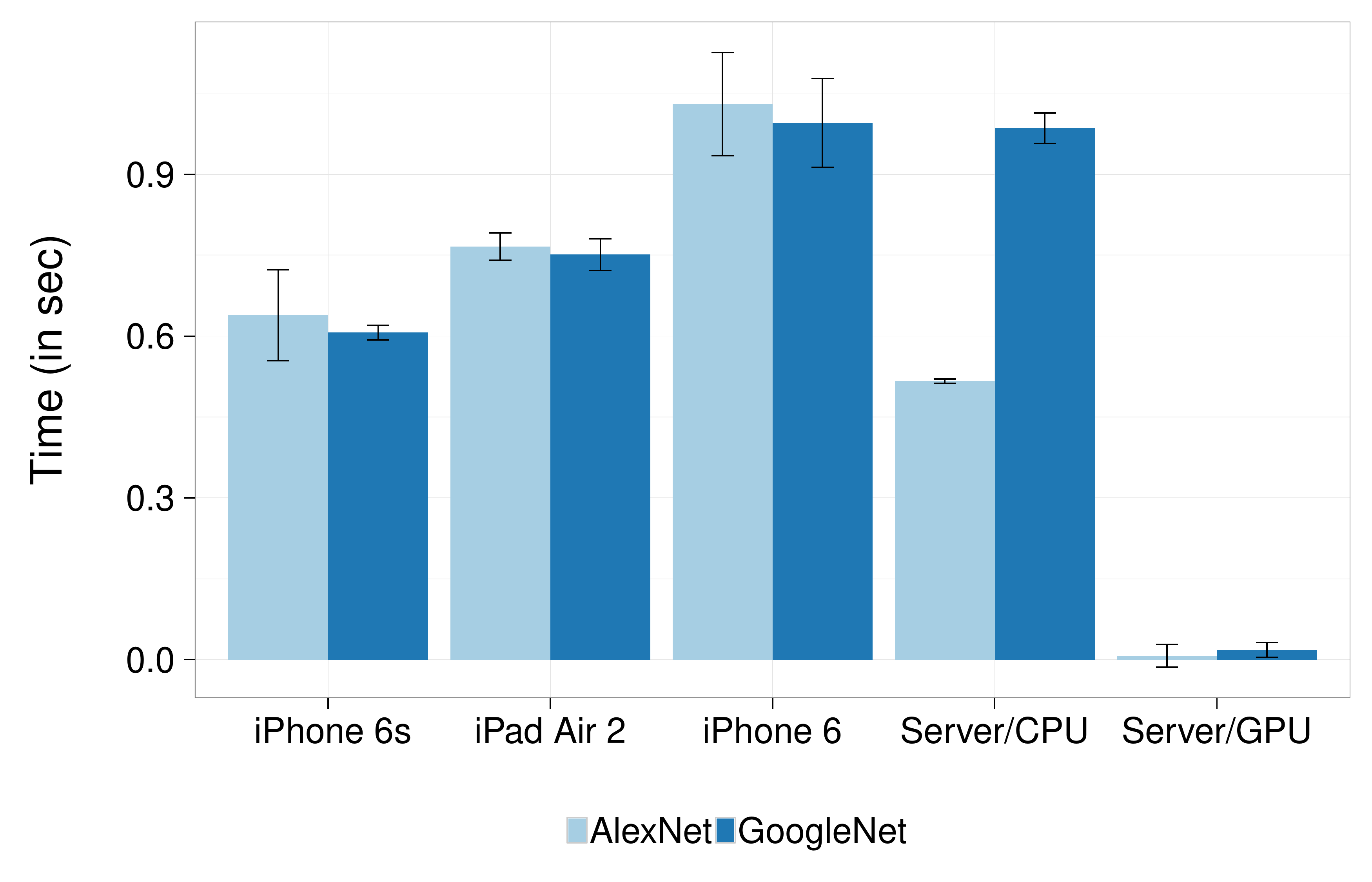}
  \caption{\textbf{AlexNet Classification Runtime on Different Devices:} Mobile Devices like current iOS devices deliver an acceptable performance. GPUs deliver the best performance.}
  \label{fig:experiments_deployment_deployment}
\end{figure}

In Figure~\ref{fig:experiments_deployment_deployment} we compare the inference
time on different platforms. Not surprisingly, the best performance is achieved
on GPUs (TitanX). The performance penalty on mobile devices is acceptable. The 
inference time on a iPad Air 2 with an A8X custom chips is on average only 
22\,\% slower than on a server side CPU. The performance of Apple's newest 
mobile CPU (A9) is only 3.7\,\% worse than the server side performance. In 
particular, the mobile deployment performance of GoogLeNet is slightly better 
than that of AlexNet.

As the performance on the mobile platform is acceptable and the object
recognition tasks has a static nature, we integrated the model into the iPad
application to give the user the opportunity to quickly verify the taken image.
In the future, we explore approaches for further optimizing networks for mobile
and embedded deployments, e.\,g., using compressing
techniques~\cite{DBLP:journals/corr/IandolaMAHDK16}.

\begin{figure}[t]
  \centering    \includegraphics[width=.49\textwidth]{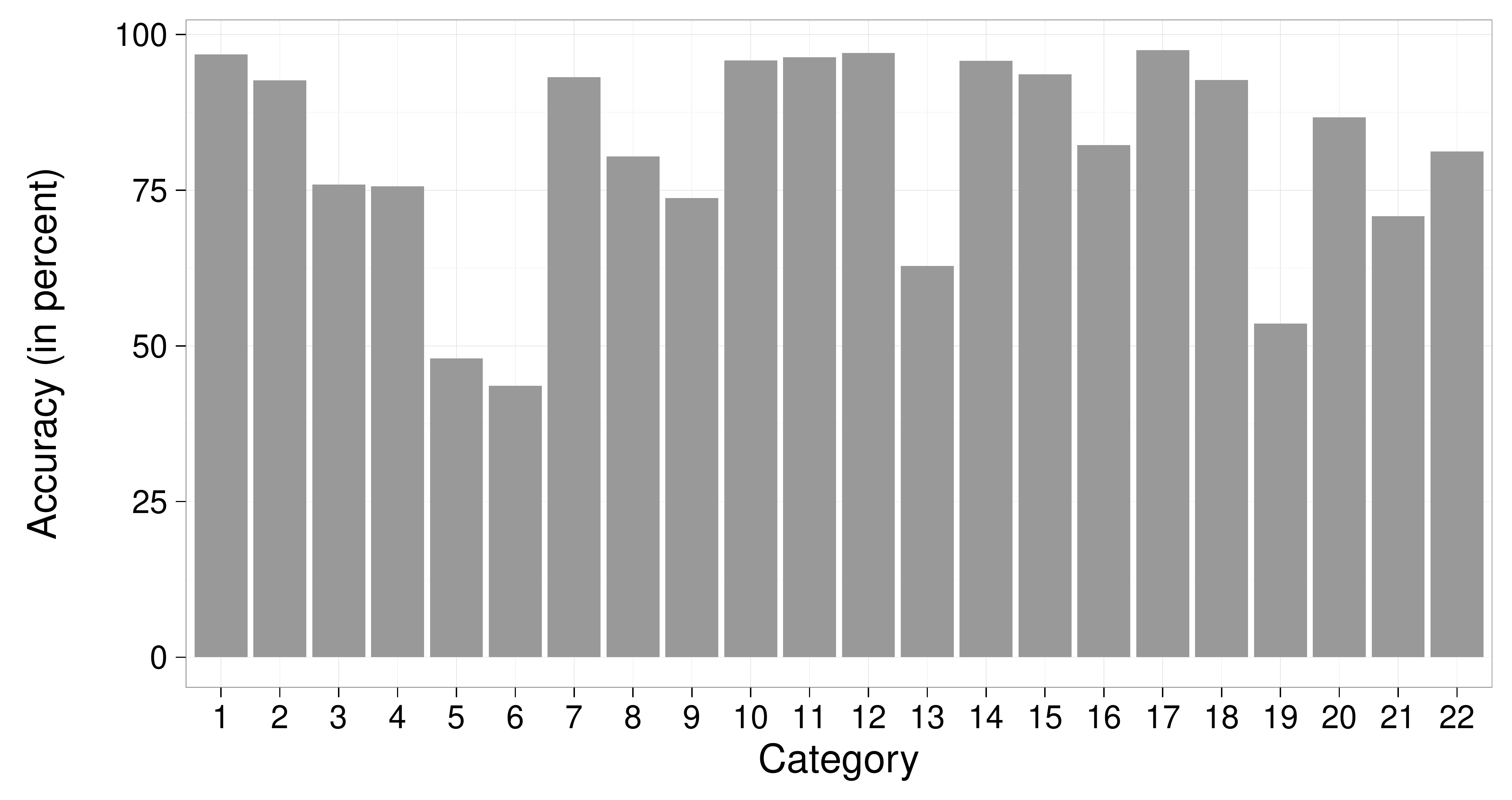}
  \caption{\textbf{Mobile Classification Accuracy in Real-World Deployment:} Accuracy varies depending on category between 44\,\% and 97\,\%. In average 81\,\% accuracy was achieved.}
  \label{fig:experiments_app_performance_classification_performance}
\end{figure}

The application was successfully deployed in production.
Figure~\ref{fig:experiments_app_performance_classification_performance} shows
the average classification performance computed using a sample of 204,883
classifications collected over a period of multiple weeks.  As previously 
described the classification is done within the mobile application after the 
image has been taken, i.\,e.\ the CNN has not seen the data before. In contrast 
to the training set, the data was not carefully prepared and pre-processed. The
application utilizes a reduced set of 21\,categories. As shown in the figure,
the accuracy varies between 44\,\% in category 6 to 98\,\% in category 1. In
average we were able to achieve an accuracy of 81\,\% on data scored in
real-time within the mobile application. In the future, we will utilize the new 
data to improve the accuracy in the low-performing categories.

\subsection{Social Media Analytics}
\label{sec:implementation_MarketingAnalytics}

In the following with utilize a CNN for recognition of vehicle models in social
media data collected from Twitter. A Python application was developed to
display the currently streaming image with its top five classifications
predicted by the neural network. Further experiments were conducted using focus
regions within the image to improve classification accuracy. More details are
discussed in the following sections.

The Cars dataset released by the Stanford AI Lab~\cite{carsdataset} consists of
16,185 images grouped into 196 categories of the form: Make, Model, Year. We
decreased the granularity of the classes into 49 separate car brands as we were 
primarily concerned with detecting different brands. We used a
pre-trained ImageNet GoogLeNet model from the Berkeley Vision and Learning
Center (BVLC)~\cite{bvlc-googlenet}. We then applied transfer learning
techniques to further train our model on a car models dataset.

To process social media data, we implemented a two version: (i) the standard
version processes the image is processed in its original form, (ii) the
region-search version adds an additional pre-processing step: First, we conduct
a selective search~\cite{UijlingsIJCV2013} on the image to isolate object
regions within the image. Next, these regions are passed to an ILSVRC13
detection network provided by BVLC~\cite{bvlc-rcnn} in order to extract object
regions containing cars. Then, these extracted car regions are passed to our
model for inferencing. Finally, the top 5 most confident class predictions over
all car regions are selected for classification of the input image.

\begin{figure}[t]
  \centering
    \includegraphics[width=.4\textwidth]{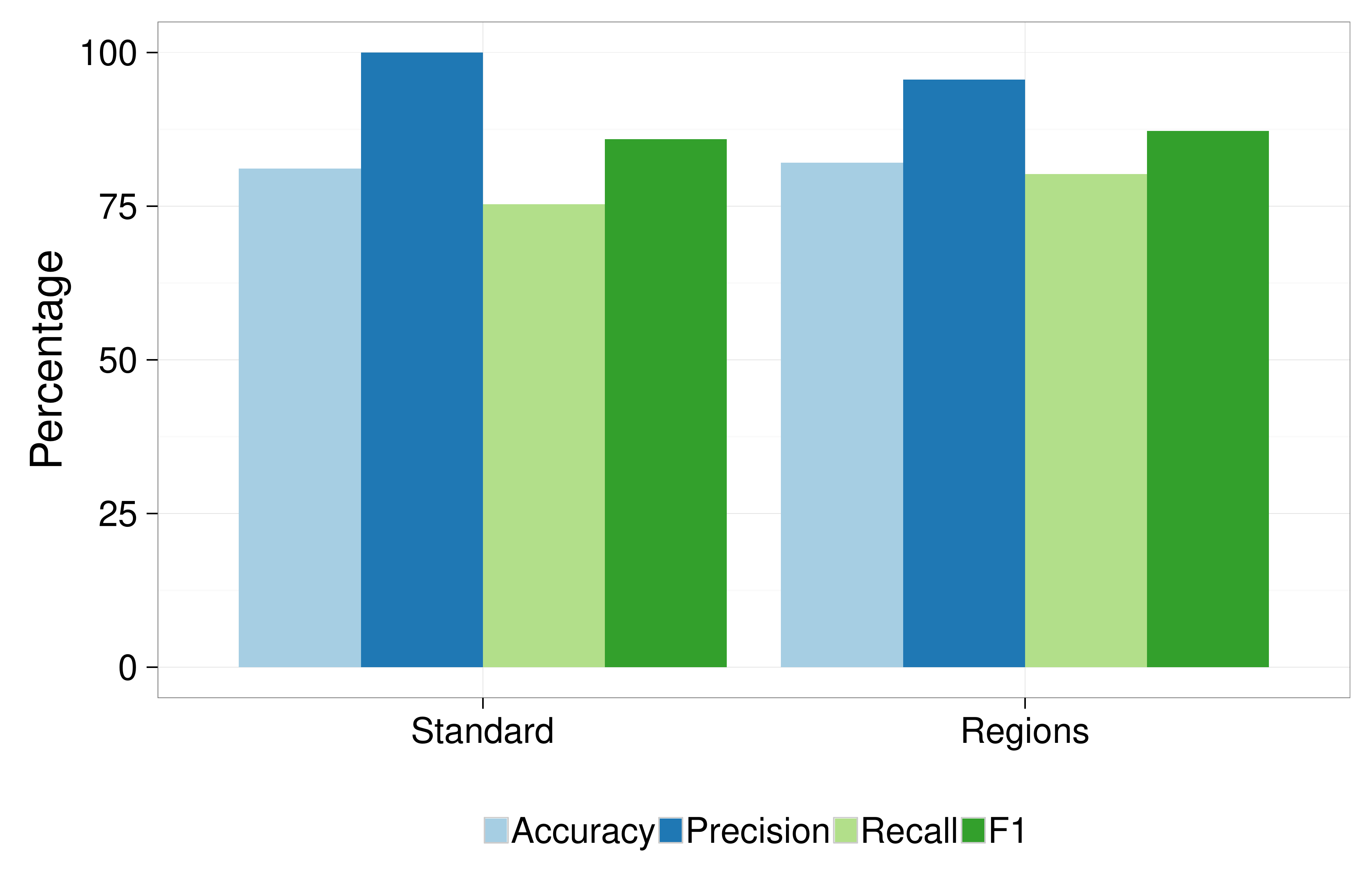}
  \caption{Social Media Analytics: Top-5 Performance using the standard and 
  		   search-search versions.}
  \label{fig:marketing_analytics_top5}
\end{figure}

\begin{figure}[t]
  \centering
    \includegraphics[width=.4\textwidth]{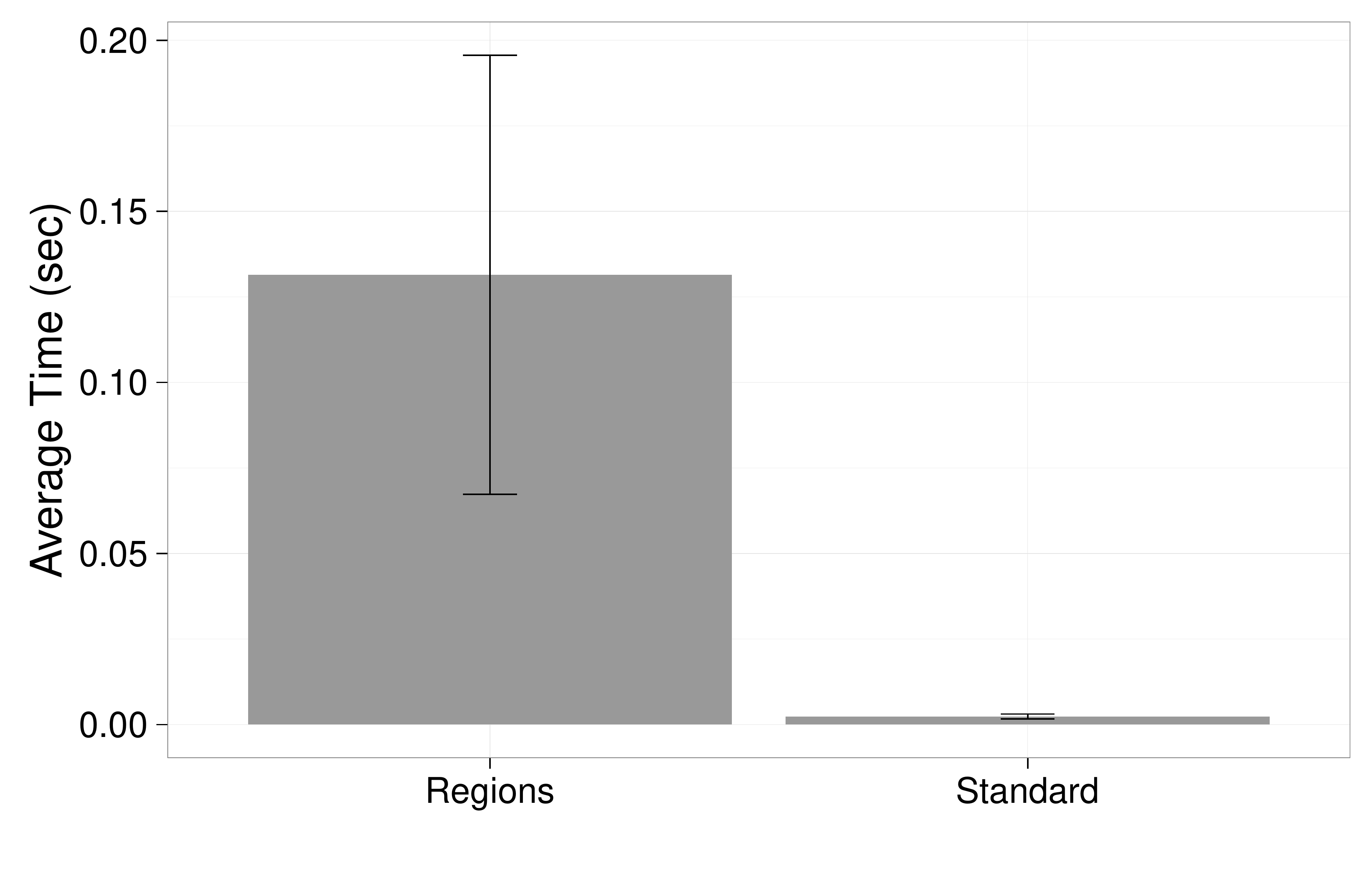}
  \caption{Social Media Analytics Inference Times for standard and 
  		   region-search version}
  \label{fig:marketing_analytics_time}
\end{figure}

We used a sample of 106 images from the Twitter feed to measure our model's
performance in five categories: classification accuracy, precision, recall,
F1 score, and processing speed per image.
Figures~\ref{fig:marketing_analytics_top5}
and~\ref{fig:marketing_analytics_time} show a comparison of the performance
metrics between our the standard (i) and region-search version (ii).

Figure~\ref{fig:marketing_analytics_top5} compare both models in terms of their
classification performance for the top-5 predicted classes. For the standard
workflow, we observed a top-5 accuracy of 81.1\,\% and F1 score of 85.9\,\%.
With the region-search version (ii), the top-5 accuracy improved to 82.1\,\%
and the F1 score to 87.2\,\%. This is only a very modest, statically
insignificant increase of $\sim$1\,\% . However, we also measured our
region-based workflow against only images which our standard version failed to
predict correctly, which lead to an improvement in the top-5 accuracy of
53.1\,\%.

Figure~\ref{fig:marketing_analytics_time} compares both models in terms of
processing speed in seconds per image. We found that our standard workflow
processed each image on average 0.002 seconds. The standard version 
significantly outperforms the region-search version, which took an average of 
0.13 seconds/image. This outcome is expected due to the extra image 
preprocessing steps involved in the region-search version.

Overall, we found that both workflows performed the same over the sampled
images. However, the region-based workflow showed significant improvement in
images where the standard workflow failed, specifically in images where the car
being analyzed did not  encompass the bulk of the
image. Our region-based approach was able to
better identify a focus region in the image to pass to our classifier,
resulting in more accurate predictions on such images.

\section{Conclusion and On-Going Research}

Deep learning enables computers to learn objects and representations, it is
however, associated with several challenges: it requires massive amounts of
data, new tools and infrastructures for computation and data. We showed that
existing model architectures and transfer learning can be applied to solve
computer vision problems in the automotive domain. In this paper, we showed the
successful deployment of deep learning for visual inspection and social media
analytics. We successfully showed the trade-offs when training and deploying
deep neural networks on a diverse set of environments (on-premise, cloud). We showed the effectiveness of the training classifier achieving an accuracy of 85\,\% during real-world use.

Several challenges for a broader deployment of deep learning remain: The
availability of labeled data is critical for development and refinement of deep
learning systems. Unfortunately, the datasets publicly available (other than
ImageNet) are not sufficient for advanced systems, e.\,g.\ for autonomous
driving. Curating training data beyond existing public datasets is a tedious
task and requires significant effort. To improve the speed of innovation, the
training time needs to be further improved.

In the future, we will: (i) investigate distributed deep learning systems to
improve training times for more complex networks and larger data sets, (ii)
assess and curate available datasets for computer vision use cases in the
domain of autonomous driving and (iii) evaluate natural understanding deep
learning models (e.\,g., sequence-to-sequence learning).

\vspace{3mm}
\subsubsection*{Acknowledgements} We thank Ken Kennedy and Colan Biemer for proof-reading. We acknowledge Darius Cepulis for his early work on deep learning benchmarks.


\footnotesize
\begin{thebibliography}{10}

\bibitem{Bengio-et-al-2015-Book}
Yoshua Bengio, Ian~J. Goodfellow, and Aaron Courville.
\newblock Deep learning.
\newblock Book in preparation for MIT Press, 2015.

\bibitem{eosl}
Trevor~J. Hastie, Robert~John Tibshirani, and Jerome~H. Friedman.
\newblock {\em The elements of statistical learning: data mining, inference,
  and prediction}.
\newblock Springer series in statistics. Springer, New York, 2009.

\bibitem{bd-workloads-infrastructure-2015}
Andre Luckow, Ken Kennedy, Fabian Manhardt, Emil Djerekarov, Bennie Vorster,
  and Amy Apon.
\newblock Automotive big data: Applications, workloads and infrastructures.
\newblock In {\em {Proceedings of IEEE Conference on Big Data}}, Santa Clara,
  CA, USA, 2015. IEEE.

\bibitem{2015arXiv150401716H}
B.~{Huval}, T.~{Wang}, S.~{Tandon}, J.~{Kiske}, W.~{Song}, J.~{Pazhayampallil},
  M.~{Andriluka}, P.~{Rajpurkar}, T.~{Migimatsu}, R.~{Cheng-Yue}, F.~{Mujica},
  A.~{Coates}, and A.~Y. {Ng}.
\newblock {An Empirical Evaluation of Deep Learning on Highway Driving}.
\newblock {\em ArXiv e-prints}, April 2015.

\bibitem{conf/nips/Pomerleau90}
Dean Pomerleau.
\newblock Rapidly adapting artificial neural networks for autonomous
  navigation.
\newblock In Richard Lippmann, John~E. Moody, and David~S. Touretzky, editors,
  {\em NIPS}, pages 429--435. Morgan Kaufmann, 1990.

\bibitem{schmidhuber}
J.~Schmidhuber.
\newblock Deep learning in neural networks: An overview.
\newblock {\em Neural Networks}, 61:85--117, 2015.
\newblock Published online 2014; based on TR arXiv:1404.7828 [cs.NE].

\bibitem{NIPS2012_4824}
Alex Krizhevsky, Ilya Sutskever, and Geoffrey~E. Hinton.
\newblock Imagenet classification with deep convolutional neural networks.
\newblock In F.~Pereira, C.J.C. Burges, L.~Bottou, and K.Q. Weinberger,
  editors, {\em Advances in Neural Information Processing Systems 25}, pages
  1097--1105. Curran Associates, Inc., 2012.

\bibitem{hinton2012deep}
Geoffrey Hinton, Li~Deng, Dong Yu, George~E Dahl, Abdel-rahman Mohamed, Navdeep
  Jaitly, Andrew Senior, Vincent Vanhoucke, Patrick Nguyen, Tara~N Sainath,
  et~al.
\newblock Deep neural networks for acoustic modeling in speech recognition: The
  shared views of four research groups.
\newblock {\em Signal Processing Magazine, IEEE}, 29(6):82--97, 2012.

\bibitem{DBLP:journals/corr/abs-1206-5538}
Yoshua Bengio, Aaron~C. Courville, and Pascal Vincent.
\newblock Unsupervised feature learning and deep learning: {A} review and new
  perspectives.
\newblock {\em CoRR}, abs/1206.5538, 2012.

\bibitem{HinSal06}
Geoffrey Hinton and Ruslan Salakhutdinov.
\newblock Reducing the dimensionality of data with neural networks.
\newblock {\em Science}, 313(5786):504 -- 507, 2006.

\bibitem{DBLP:journals/corr/RussakovskyDSKSMHKKBBF14}
Olga Russakovsky, Jia Deng, Hao Su, Jonathan Krause, Sanjeev Satheesh, Sean Ma,
  Zhiheng Huang, Andrej Karpathy, Aditya Khosla, Michael~S. Bernstein,
  Alexander~C. Berg, and Fei{-}Fei Li.
\newblock Imagenet large scale visual recognition challenge.
\newblock {\em CoRR}, abs/1409.0575, 2014.

\bibitem{ms_deepresidlearning_2015arXiv}
K.~{He}, X.~{Zhang}, S.~{Ren}, and J.~{Sun}.
\newblock {Deep Residual Learning for Image Recognition}.
\newblock {\em ArXiv e-prints}, December 2015.

\bibitem{alphago}
David Silver, Aja Huang, Chris~J. Maddison, Arthur Guez, Laurent Sifre, George
  van~den Driessche, Julian Schrittwieser, Ioannis Antonoglou, Veda
  Panneershelvam, Marc Lanctot, Sander Dieleman, Dominik Grewe, John Nham, Nal
  Kalchbrenner, Ilya Sutskever, Timothy Lillicrap, Madeleine Leach, Koray
  Kavukcuoglu, Thore Graepel, and Demis Hassabis.
\newblock Mastering the game of go with deep neural networks and tree search.
\newblock {\em Nature}, 529(7587):484--489, 01 2016.

\bibitem{DBLP:journals/corr/VasilacheJMCPL14}
Nicolas Vasilache, Jeff Johnson, Micha{\"{e}}l Mathieu, Soumith Chintala,
  Serkan Piantino, and Yann LeCun.
\newblock Fast convolutional nets with fbfft: {A} {GPU} performance evaluation.
\newblock {\em CoRR}, abs/1412.7580, 2014.

\bibitem{cudnn}
{NVIDIA cuDNN}.
\newblock \url{https://developer.nvidia.com/cuDNN}, 2015.

\bibitem{dl_mkl}
{{Gennady Fedorov and Vadim Pirogov and Nikita Shustrov}}.
\newblock {Deep Neural Network Technical Preview for Intel Math Kernel Library
  (Intel MKL)}.
\newblock \url{http://intel.ly/1RRx9L2}, 2015.

\bibitem{scikit-learn}
F.~Pedregosa, G.~Varoquaux, A.~Gramfort, V.~Michel, B.~Thirion, O.~Grisel,
  M.~Blondel, P.~Prettenhofer, R.~Weiss, V.~Dubourg, J.~Vanderplas, A.~Passos,
  D.~Cournapeau, M.~Brucher, M.~Perrot, and E.~Duchesnay.
\newblock Scikit-learn: Machine learning in {P}ython.
\newblock {\em Journal of Machine Learning Research}, 12:2825--2830, 2011.

\bibitem{pylearn2_arxiv_2013}
Ian~J. Goodfellow, David Warde-Farley, Pascal Lamblin, Vincent Dumoulin, Mehdi
  Mirza, Razvan Pascanu, James Bergstra, Fr{\'{e}}d{\'{e}}ric Bastien, and
  Yoshua Bengio.
\newblock Pylearn2: a machine learning research library.
\newblock {\em arXiv preprint arXiv:1308.4214}, 2013.

\bibitem{dato}
{Danny Bickson}.
\newblock {Dato's Deep Learning Toolkit}.
\newblock \url{http://blog.dato.com/deep-learning-blog-post}, 2015.

\bibitem{dl4j}
{Deep Learning for Java}.
\newblock \url{http://deeplearning4j.org/}, 2015.

\bibitem{RJournal_2010-1_Guenther+Fritsch}
Frauke G\"unther and Stefan Fritsch.
\newblock { Neuralnet: Training of neural networks }.
\newblock {\em {The R Journal}}, 2(1):30--38, jun 2010.

\bibitem{DBLP:journals/corr/JiaSDKLGGD14}
Yangqing Jia, Evan Shelhamer, Jeff Donahue, Sergey Karayev, Jonathan Long,
  Ross~B. Girshick, Sergio Guadarrama, and Trevor Darrell.
\newblock Caffe: Convolutional architecture for fast feature embedding.
\newblock {\em CoRR}, abs/1408.5093, 2014.

\bibitem{tensorflow2015-whitepaper}
Mart\'{\i}n Abadi, Ashish Agarwal, Paul Barham, Eugene Brevdo, Zhifeng Chen,
  Craig Citro, Greg~S. Corrado, Andy Davis, Jeffrey Dean, Matthieu Devin,
  Sanjay Ghemawat, Ian Goodfellow, Andrew Harp, Geoffrey Irving, Michael Isard,
  Yangqing Jia, Rafal Jozefowicz, Lukasz Kaiser, Manjunath Kudlur, Josh
  Levenberg, Dan Man\'{e}, Rajat Monga, Sherry Moore, Derek Murray, Chris Olah,
  Mike Schuster, Jonathon Shlens, Benoit Steiner, Ilya Sutskever, Kunal Talwar,
  Paul Tucker, Vincent Vanhoucke, Vijay Vasudevan, Fernanda Vi\'{e}gas, Oriol
  Vinyals, Pete Warden, Martin Wattenberg, Martin Wicke, Yuan Yu, and Xiaoqiang
  Zheng.
\newblock {TensorFlow}: Large-scale machine learning on heterogeneous systems,
  2015.
\newblock Software available from tensorflow.org.

\bibitem{cntk}
Dong Yu, Adam Eversole, Mike Seltzer, Kaisheng Yao, Oleksii Kuchaiev, Yu~Zhang,
  Frank Seide, Zhiheng Huang, Brian Guenter, Huaming Wang, Jasha Droppo,
  Geoffrey Zweig, Chris Rossbach, Jie Gao, Andreas Stolcke, Jon Currey, Malcolm
  Slaney, Guoguo Chen, Amit Agarwal, Chris Basoglu, Marko Padmilac, Alexey
  Kamenev, Vladimir Ivanov, Scott Cypher, Hari Parthasarathi, Bhaskar Mitra,
  Baolin Peng, and Xuedong Huang.
\newblock An introduction to computational networks and the computational
  network toolkit.
\newblock Technical report, October 2014.

\bibitem{dsstne}
Amazon.
\newblock {Deep Scalable Sparse Tensor Network Engine (DSSTNE) }.
\newblock \url{https://github.com/amznlabs/amazon-dsstne}, 2016.

\bibitem{DBLP:journals/corr/ChenLLLWWXXZZ15}
Tianqi Chen, Mu~Li, Yutian Li, Min Lin, Naiyan Wang, Minjie Wang, Tianjun Xiao,
  Bing Xu, Chiyuan Zhang, and Zheng Zhang.
\newblock Mxnet: {A} flexible and efficient machine learning library for
  heterogeneous distributed systems.
\newblock {\em CoRR}, abs/1512.01274, 2015.

\bibitem{collobert2011torch7}
Ronan Collobert, Koray Kavukcuoglu, and Cl{\'e}ment Farabet.
\newblock Torch7: A matlab-like environment for machine learning.
\newblock In {\em BigLearn, NIPS Workshop}, number EPFL-CONF-192376, 2011.

\bibitem{paddle}
Baidu.
\newblock Paddlepaddle.
\newblock {\url{http://www.paddlepaddle.org/}}, 2016.

\bibitem{digits}
{{NVIDIA}}.
\newblock {DIGITS}.
\newblock \url{https://developer.nvidia.com/digits}, 2016.

\bibitem{keras}
{{François Chollet et.\,al}}.
\newblock {Keras: Deep Learning library for Theano and TensorFlow}.
\newblock \url{http://keras.io/}, 2016.

\bibitem{lasagne}
{{Jan Schlüter et.\,al}}.
\newblock {Lasagne: Neural Network Tools for Theano}.
\newblock \url{https://github.com/Lasagne/Lasagne}, 2016.

\bibitem{fpga-dnn}
Kalin Ovtcharov, Olatunji Ruwase, Joo-Young Kim, Jeremy Fowers, Karin Strauss,
  and Eric~S. Chung.
\newblock Accelerating deep convolutional neural networks using specialized
  hardware, February 2015.

\bibitem{40565}
Jeffrey Dean, Greg~S. Corrado, Rajat Monga, Kai Chen, Matthieu Devin, Quoc~V.
  Le, Mark~Z. Mao, Marc’Aurelio Ranzato, Andrew Senior, Paul Tucker, Ke~Yang,
  and Andrew~Y. Ng.
\newblock Large scale distributed deep networks.
\newblock In {\em NIPS}, 2012.

\bibitem{DBLP:journals/corr/WuYSDS15}
Ren Wu, Shengen Yan, Yi~Shan, Qingqing Dang, and Gang Sun.
\newblock Deep image: Scaling up image recognition.
\newblock {\em CoRR}, abs/1501.02876, 2015.

\bibitem{Xing:179}
Eric~P. Xing, Qirong Ho, Pengtao Xie, and Dai Wei.
\newblock Strategies and principles of distributed machine learning on big
  data.
\newblock {\em Engineering}, 2(2):179, 2016.

\bibitem{2011arXiv1106.5730N}
F.~{Niu}, B.~{Recht}, C.~{Re}, and S.~J. {Wright}.
\newblock {HOGWILD!: A Lock-Free Approach to Parallelizing Stochastic Gradient
  Descent}.
\newblock {\em ArXiv e-prints}, June 2011.

\bibitem{hadoop}
{Hadoop: Open Source Implementation of MapReduce}.
\newblock http://hadoop.apache.org/.

\bibitem{Zaharia:2010:SCC:1863103.1863113}
Matei Zaharia, Mosharaf Chowdhury, Michael~J. Franklin, Scott Shenker, and Ion
  Stoica.
\newblock Spark: Cluster computing with working sets.
\newblock In {\em Proceedings of the 2Nd USENIX Conference on Hot Topics in
  Cloud Computing}, HotCloud'10, pages 10--10, Berkeley, CA, USA, 2010. USENIX
  Association.

\bibitem{spark_mlp}
Alexander Ulanov.
\newblock {Spark Multilayer perceptron classifier}.
\newblock
  \url{https://spark.apache.org/docs/latest/ml-classification-regression.html#multilayer-perceptron-classifier},\url{https://issues.apache.org/jira/browse/SPARK-5575},
  2016.

\bibitem{mllib}
Mllib.
\newblock \url{https://spark.apache.org/mllib/}, 2014.

\bibitem{caffeOnSpark}
Cyprien Noel, Jun Shi, and Andy Feng.
\newblock {Large Scale Distributed Deep Learning on Hadoop Clusters}.
\newblock
  \url{http://yahoohadoop.tumblr.com/post/129872361846/large-scale-distributed-deep-learning-on-hadoop},
  2016.

\bibitem{2015arXiv151106051M}
P.~{Moritz}, R.~{Nishihara}, I.~{Stoica}, and M.~I. {Jordan}.
\newblock {SparkNet: Training Deep Networks in Spark}.
\newblock {\em ArXiv e-prints: \url{http://arxiv.org/abs/1511.06051}}, November
  2015.

\bibitem{tensorspark}
Christopher Smith, Ushnish De, and Christopher Nguyen.
\newblock {Distributed TensorFlow: Scaling Google’s Deep Learning Library on
  Spark}.
\newblock
  \url{https://arimo.com/machine-learning/deep-learning/2016/arimo-distributed-tensorflow-on-spark/},
  2016.

\bibitem{1_bit_sgd}
Frank Seide, Hao Fu, Jasha Droppo, Gang Li, and Dong Yu.
\newblock 1-bit stochastic gradient descent and application to data-parallel
  distributed training of speech dnns.
\newblock September 2014.

\bibitem{DBLP:journals/corr/IandolaAMK15}
Forrest~N. Iandola, Khalid Ashraf, Matthew~W. Moskewicz, and Kurt Keutzer.
\newblock Firecaffe: near-linear acceleration of deep neural network training
  on compute clusters.
\newblock {\em CoRR}, abs/1511.00175, 2015.

\bibitem{DBLP:journals/corr/VishnuSD16}
Abhinav Vishnu, Charles Siegel, and Jeffrey Daily.
\newblock Distributed tensorflow with {MPI}.
\newblock {\em CoRR}, abs/1603.02339, 2016.

\bibitem{arimo_tf}
Christopher Smith, Christopher Nguyen, and Ushnish De.
\newblock {Distributed TensorFlow: Scaling Google’s Deep Learning Library on
  Spark}.
\newblock
  \url{https://arimo.com/machine-learning/deep-learning/2016/arimo-distributed-tensorflow-on-spark/},
  2016.

\bibitem{amazon_gpu}
Jeff Barr.
\newblock New g2 instance type with 4x more gpu power.
\newblock
  \url{https://aws.amazon.com/blogs/aws/new-g2-instance-type-with-4x-more-gpu-power/},
  2015.

\bibitem{google_cml}
{{Google}}.
\newblock {Cloud Machine Learning}.
\newblock \url{https://cloud.google.com/ml/}, 2016.

\bibitem{emr}
{{Amazon Web Services}}.
\newblock {{Elastic Map Reduce Service}}.
\newblock \url{http://aws.amazon.com/de/elasticmapreduce/}, 2013.

\bibitem{azure-hdi}
Microsoft.
\newblock {HDInsight}.
\newblock \url{https://azure.microsoft.com/de-de/services/hdinsight/}, 2016.

\bibitem{google-dataproc}
Google.
\newblock {Cloud Dataproc}.
\newblock \url{https://cloud.google.com/dataproc/}, 2016.

\bibitem{google_prediction}
{Google}.
\newblock {Prediction API}.
\newblock \url{https://cloud.google.com/prediction/}, 2015.

\bibitem{azureml}
{Azure ML}.
\newblock \url{http://azureml.net/}, 2015.

\bibitem{amazonml}
{Amazon Machine Learning}.
\newblock \url{https://aws.amazon.com/machine-learning/}, 2015.

\bibitem{project_oxford}
Microsoft.
\newblock {Project Oxford}.
\newblock \url{http://www.projectoxford.ai/}, 2015.

\bibitem{google_vision}
{{Google}}.
\newblock {Cloud Vision API}.
\newblock \url{https://cloud.google.com/vision/}, 2016.

\bibitem{google_nlp}
{{Google}}.
\newblock {Cloud Natural Language API}.
\newblock \url{https://cloud.google.com/natural-language/}, 2016.

\bibitem{ibm_watson}
{{IBM}}.
\newblock {Watson Developer Cloud}.
\newblock
  \url{http://www.ibm.com/smarterplanet/us/en/ibmwatson/developercloud}, 2016.

\bibitem{alexnet_tf}
{Tensorflow AlexNet}.
\newblock
  \url{https://github.com/tensorflow/tensorflow/blob/master/tensorflow/models/image/alexnet/alexnet_benchmark.py},
  2016.

\bibitem{Houben-IJCNN-2013}
Sebastian Houben, Johannes Stallkamp, Jan Salmen, Marc Schlipsing, and
  Christian Igel.
\newblock Detection of traffic signs in real-world images: The {G}erman
  {T}raffic {S}ign {D}etection {B}enchmark.
\newblock In {\em International Joint Conference on Neural Networks}, number
  1288, 2013.

\bibitem{Places}
Bolei Zhou, Agata Lapedriza, Jianxiong Xiao, Antonio Torralba, and Aude Oliva.
\newblock Learning deep features for scene recognition using places database.
\newblock 2014.

\bibitem{Geiger2012CVPR}
Andreas Geiger, Philip Lenz, and Raquel Urtasun.
\newblock Are we ready for autonomous driving? the kitti vision benchmark
  suite.
\newblock In {\em Conference on Computer Vision and Pattern Recognition
  (CVPR)}, 2012.

\bibitem{krause20133d}
Jonathan Krause, Michael Stark, Jia Deng, and Li~Fei-Fei.
\newblock 3d object representations for fine-grained categorization.
\newblock In {\em Proceedings of the IEEE International Conference on Computer
  Vision Workshops}, pages 554--561, 2013.

\bibitem{luckow-2015}
Andre Luckow, Ken Kennedy, Fabian Manhardt, Emil Djerekarov, Bennie Vorster,
  and Amy Apon.
\newblock Automotive big data: Applications, workloads and infrastructures.
\newblock In {\em Big Data (Big Data), 2015 IEEE International Conference on},
  pages 1201--1210, Oct 2015.

\bibitem{DBLP:journals/corr/SzegedyLJSRAEVR14}
Christian Szegedy, Wei Liu, Yangqing Jia, Pierre Sermanet, Scott Reed, Dragomir
  Anguelov, Dumitru Erhan, Vincent Vanhoucke, and Andrew Rabinovich.
\newblock Going deeper with convolutions.
\newblock {\em CoRR}, abs/1409.4842, 2014.

\bibitem{googlenet-v2}
Sergey Ioffe and Christian Szegedy.
\newblock Batch normalization: Accelerating deep network training by reducing
  internal covariate shift.
\newblock {\em CoRR}, abs/1502.03167, 2015.

\bibitem{vgg_very_deep_conv}
Karen Simonyan and Andrew Zisserman.
\newblock Very deep convolutional networks for large-scale image recognition.
\newblock {\em CoRR}, abs/1409.1556, 2014.

\bibitem{google_inception_2015arXiv}
C.~{Szegedy}, V.~{Vanhoucke}, S.~{Ioffe}, J.~{Shlens}, and Z.~{Wojna}.
\newblock {Rethinking the Inception Architecture for Computer Vision}.
\newblock {\em ArXiv e-prints}, December 2015.

\bibitem{DBLP:journals/corr/IandolaMAHDK16}
Forrest~N. Iandola, Matthew~W. Moskewicz, Khalid Ashraf, Song Han, William~J.
  Dally, and Kurt Keutzer.
\newblock Squeezenet: Alexnet-level accuracy with 50x fewer parameters and
  {\textless}1mb model size.
\newblock {\em CoRR}, abs/1602.07360, 2016.

\bibitem{carsdataset}
Jonathan Krause, Michael Stark, Jia Deng, and Li~Fei-Fei.
\newblock 3d object representations for fine-grained categorization.
\newblock In {\em 4th IEEE Workshop on 3D Representation and Recognition, at
  ICCV 2013}. IEEE, 2013.

\bibitem{bvlc-googlenet}
Berkeley Vision and Learning Center.
\newblock Bvlc googlenet model.
\newblock
  \url{https://github.com/BVLC/caffe/tree/master/models/bvlc_googlenet}, 2015.

\bibitem{UijlingsIJCV2013}
J.~R.~R. Uijlings, K.~E.~A. van~de Sande, T.~Gevers, and A.~W.~M. Smeulders.
\newblock Selective search for object recognition.
\newblock {\em International Journal of Computer Vision}, 104(2):154--171,
  2013.

\bibitem{bvlc-rcnn}
Berkeley Vision and Learning Center.
\newblock Bvlc reference rcnn ilsvrc13 model.
\newblock
  \url{https://github.com/BVLC/caffe/tree/master/models/bvlc_reference_rcnn_ilsvrc13},
  2015.

\end{thebibliography}
\end{document}